\documentclass{article}
\usepackage{aaai17}
\usepackage{times}
\usepackage{helvet}
\usepackage{courier}
\usepackage{url}            
\usepackage{booktabs}       
\usepackage{amsfonts}       
\usepackage{nicefrac}       
\usepackage{microtype}      
\usepackage{appendix}

\usepackage{amsmath}             
\usepackage{bbm}                 
\usepackage[psamsfonts]{amssymb} 
\usepackage{graphicx}            
\usepackage{multirow}
\usepackage{dcolumn}
\usepackage{cleveref}
\usepackage{tabularx}
\usepackage{siunitx}
\newcommand{\citet}[1]{\citeauthor{#1}~\shortcite{#1}}

\DeclareFontFamily{OT1}{mathc}{}
\DeclareFontShape{OT1}{mathc}{m}{n}{ <-> mathc10 }{}
\DeclareRobustCommand\xmcal[1]{\text{\usefont{OT1}{mathc}{m}{n}#1}}
\renewcommand{\mathcal}{\xmcal}

\newcommand{\gaussian}{\mathcal{N}}

\newcommand{\vm}{\mathcal{vM}}

\newcommand{\gvm}{\mathcal{GvM}}

\newcommand{\mgvm}{\mathcal{mGvM}}

\newcommand{\mvm}{\mathcal{mvM}}

\newcommand{\gp}{\mathcal{GP}}

\newcommand{\yy}{y}

\newcommand{\p}{p}

\newcommand{\q}{q}

\newcommand{\expect}[1]{\left\langle {#1} \right\rangle}

\newcommand{\free}{\mathcal{F}}

\newcommand{\entropy}{\mathcal{H}}

\newcommand{\const}{\text{const.}}

\newcommand{\params}{\theta}

\newcommand{\reals}{\mathbb{R}}

\newcommand{\integers}{\mathbb{Z}}

\newcommand{\idx}[1]{{#1}}

\newcommand{\MM}{M}

\newcommand{\DD}{D}

\newcommand{\dd}{\idx{d}}

\newcommand{\jj}{\idx{j}}

\newcommand{\mat}[1]{\boldsymbol{\uppercase{#1}}}

\newcommand{\vt}[1]{\boldsymbol{#1}}
\renewcommand{\vec}[1]{\boldsymbol{#1}}

\newcommand{\id}{I}

\newcommand{\del}{\text{d}}

\newcommand{\vyy}{\vt{\yy}}

\newcommand{\xx}{x}

\newcommand{\vxx}{\vt{x}}

\newcommand{\error}{\epsilon}

\newcommand{\verror}{\vt{\epsilon}}

\newcommand{\stdev}{\sigma}

\newcommand{\cov}{\vt{\Sigma}}

\newcommand{\mean}{\mu}

\newcommand{\vmean}{\vt{\mu}}

\newcommand{\matA}{\mat{A}}

\newcommand{\matB}{\mat{B}}

\newcommand{\matW}{\mat{W}}

\newcommand{\matK}{\mat{K}}

\newcommand{\arv}{\phi}

\newcommand{\varv}{\vt{\arv}}

\newcommand{\cmean}{\nu}

\newcommand{\vcmean}{\vt{\nu}}

\newcommand{\cstd}{\kappa}

\newcommand{\vcstd}{\vt{\kappa}}

\newcommand{\E}[1]{\cdot 10^{#1}}
\title{The Multivariate Generalised von Mises Distribution: Inference and Applications}
\author{
  Alexandre K.~W.~Navarro \\
  Department of Engineering\\
  University of Cambridge\\
  Cambridge, UK \\
  \texttt{akwn2@cam.ac.uk} \\
  \And
  Jes Frellsen \\
  Department of Engineering\\
  University of Cambridge\\
  Cambridge, UK \\
  \texttt{jf519@cam.ac.uk} \\
  \And
  Richard E.~Turner \\
  Department of Engineering\\
  University of Cambridge\\
  Cambridge, UK \\
  \texttt{ret26@cam.ac.uk} \\
}

\begin{document}
\maketitle

\begin{abstract}
Circular variables arise in a multitude of data-modelling contexts ranging from robotics to the social sciences, but they have been largely overlooked by the machine learning community. This paper partially redresses this imbalance by extending some standard probabilistic modelling tools to the circular domain. First we introduce a new multivariate distribution over circular variables, called the multivariate Generalised von Mises (mGvM) distribution. This distribution can be constructed by restricting and renormalising a general multivariate Gaussian distribution to the unit hyper-torus. Previously proposed multivariate circular distributions are shown to be special cases of this construction. Second, we introduce a new probabilistic model for circular regression inspired by Gaussian Processes, and a method for probabilistic Principal Component Analysis with circular hidden variables. These models can leverage standard modelling tools (e.g.~kernel functions and automatic relevance determination).  Third, we show that the posterior distribution in these models is a mGvM distribution which enables development of an efficient variational free-energy scheme for performing approximate inference and approximate maximum-likelihood learning.
\end{abstract}

\section{Introduction}

Many data modelling problems in science and engineering involve circular variables. For example, the spatial configuration of a molecule~\cite{boomsma08,frellsen09}, robot, or the human body~\cite{chirikjian2000} can be naturally described using a set of angles. Phase variables arise in image and audio modelling scenarios~\cite{wadhwa2013}, while directional fields are also present in fluid dynamics~\cite{jona-lasinio2012},
and neuroscience~\cite{benyishai1995}. Phase-locking to periodic signals occurs in a multitude of fields ranging from biology~\cite{gao2010} to the social sciences~\cite{brunsdon_using_2006}.

It is possible, at least in principle, to model circular variables using distributional assumptions that are appropriate for variables that live in a standard Euclidean space. For example, a na\"{i}ve application might represent a circular variable in terms of its angle $\arv \in [0,2\pi)$ and use a standard distribution over this variable (presumably restricted to the valid domain). Such an approach would, however, ignore the topology of the space e.g.~that $\arv =0$ and $\arv=2 \pi$ are equivalent. Alternatively, the circular variable can be represented as a unit vector in $\reals^{2}$,  $\vt{x} = [\cos(\arv),\sin(\arv)]^{\top}$, and a standard bivariate distribution used instead. This partially alleviates the aforementioned topological problem, but standard distributions place probability mass off the unit circle which adversely affects learning, prediction and analysis. 

In order to predict and analyse circular data it is therefore key that machine learning practitioners have at their disposal a suite of bespoke modelling, inference and learning methods that are specifically designed for circular data~\cite{lebanon2005}. The fields of circular and directional statistics have provided a toolbox of this sort~\cite{mardia_directional_2000}. However, the focus has been on fairly simple and small models that are applied to small datasets enabling MCMC to be tractably deployed for approximate inference.

The goal of this paper is to extend the existing toolbox provided by statistics, by leveraging modelling and approximate inference methods from the probabilistic machine learning field. Specifically, the paper makes three technical contributions. First, in \cref{sec:mgvm} it introduces a central multivariate distribution for circular data---called the multivariate Generalised von Mises distribution---that has elegant theoretical properties and which can be combined in a plug-and-play manner with existing probabilistic models. Second, in \cref{sec:applications} it shows that this distribution arises in two novel models that are circular versions of Gaussian Process regression and probabilistic Principal Component Analysis with circular hidden variables. Third, it develops efficient approximate inference and learning techniques based on variational free-energy methods as demonstrated on four datasets in \cref{sec:results}.
\section{Circular distributions primer\label{sec:review}}
In order to explain the context and rationale behind the contributions made in this paper, it is necessary to know a little background on circular distributions. Since \emph{multidimensional} circular distributions are not generally well-known in the machine learning community, we present a brief review of the main concepts related to these distributions in this section. The expert reader can jump to \cref{sec:mgvm} where the multivariate Generalised von Mises distribution is introduced.

A univariate circular distribution is a probability distribution defined over the unit circle. Such distributions can be constructed by wrapping, marginalising or conditioning standard distributions defined in Euclidean spaces and are classified as \emph{wrapped}, \emph{projected} or \emph{intrinsic} according to the geometric interpretation of their construction.

More precisely, the \emph{wrapped} approach consists of taking a univariate distribution $\p(\xx)$ defined on the real line, parametrising any point $\xx \in \reals$ as $\xx = \arv + 2 \pi k$ with $k \in \integers$ and summing over all $k$ so that $\p(\xx)$ is wrapped around the unit circle. The most commonly used wrapped distribution is the Wrapped Gaussian distribution~\cite{ferrari_wrapping_2009,jona-lasinio2012}.

An alternative approach takes a standard bivariate distribution $\p(\xx,\yy)$ that places probability mass over $\reals^2$, transforms it to polar coordinates $[\xx, \yy]^{\top} \rightarrow [r\cos\arv, r\sin\arv]^{\top}$ and marginalises out the radial component $\int_{0}^{\infty}\p(r\cos\arv, r\sin\arv)r \del r$.
This approach can be interpreted as projecting all the probability mass that lies along a ray from the origin onto the point where it crosses the unit circle.
The most commonly used projected distribution is the Projected Gaussian~\cite{wang_directional_2013}.

Instead of marginalising the radial component, circular distributions can be constructed by conditioning it to unity, $p(\xx, \yy | \xx^2+\yy^2=1)$. This can be interpreted as restricting the original bivariate density to the unit circle and renormalising. A distribution constructed in this way is called ``intrinsic'' (to the unit circle). The construction has several elegant properties. First, the resulting distribution inherits desirable characteristics of the base distribution, such as membership of the exponential family. Second, the form of the resulting density often affords more analytical tractability than those produced by wrapping or projection. The most important intrinsic distribution is the von Mises (vM),
$p(\arv|\mu,\kappa) \propto \exp(\kappa \cos(\arv-\mu))$,
which is obtained by conditioning an isotropic bivariate Gaussian to the unit circle. The vM has two parameters, the mean $\mu \in [0,2 \pi)$ and the concentration $\kappa \in \reals^{+}$.
If the covariance matrix of the bivariate Gaussian is a general real positive definite matrix, we obtain the Generalised von Mises (GvM) distribution~\cite{gatto_generalized_2007}\footnote{To be precise, Gatto and Jammalamadaka define this to be a Generalised von Mises of order 2, but since higher-order Generalised von Mises distributions are more intractable and consequently have found fewer applications, we use the shorthand throughout.}
\begin{align}
  p(\arv) \propto \exp(\kappa_1 \cos(\arv-\mu_1) + \kappa_2 \cos( 2 (\arv-\mu_2)))\,.
  \label{eq:3}
\end{align}
The GvM has four parameters, two mean-like parameters $\mu_i \in [0,2 \pi)$ and two concentration-like parameters $\kappa_i \in \reals^{+}$ and is an exponential family distribution. The GvM is generally asymmetric. It has two modes when  $4\cstd_{2} \geq \cstd_{1}$, otherwise it has one mode except when it is a uniform distribution  $\cstd_{2} = \cstd_{1} = 0$.

The GvM is arguably more tractable than the distributions obtained by wrapping or projection as its unnormalised density takes a simple form. In comparison, the unnormalised density of the wrapped normal involves an infinite sum and that of the projected normal is complex and requires special functions. However, the normalising constant of the GvM (and its higher moments) are still complicated, containing infinite sums of modified Bessel functions~\cite{gatto_computational_2008}.

In this paper the focus will be on the extensions to vectors of dependent circular variables that lie on a (hyper-) torus (although similar methods can be applied to multivariate hyper-spherical models). An example of a multivariate distribution on the hyper-torus is the multivariate von Mises (mvM) by~\citet{mardia_multivariate_2008}
\begin{align}
  \mvm(\varv) \propto \exp\Big\{\vcstd^{\top} \cos(\varv) + \sin(\varv)^{\top}\mat{G}\sin(\varv)\Big\}\,.
\end{align}
The terms $\cos(\varv)$ and $\sin(\varv)$ denote element-wise application of sine and cosine functions to the vector $\varv$, $\vcstd$ is a element-wise positive $D$-dimensional real vector, $\vcmean$ is a $D$-dimensional vector whose entries take values on $[0, 2\pi)$, and $\mat{G}$ is a matrix whose diagonal entries are all zeros.

The mvM distribution draws its name from the its property that the one dimensional conditionals, $p(\arv_d|\varv_{\neq d})$, are von Mises distributed. As shown in the Supplementary Material, this distribution can be obtained by applying the \emph{intrinsic} construction to a $2D$-dimensional Gaussian, mapping $\vt{x}\to (r \cos\varv^{\top}, r\sin\varv^{\top})^{\top}$ and assuming its precision matrix has the form
\begin{align}
  \mat{W} = \cov^{-1} =
  \begin{bmatrix}
    \Lambda & \mat{A} \\
    \mat{A}^{\top} & \Lambda
  \end{bmatrix}
  \label{eq:sparsityMvM}
\end{align}
where $\Lambda$ is a diagonal $D$ by $D$ matrix and $\mat{A}$ is an antisymmetric matrix. Other important facts about the mvM are that it bears no simple closed analytic form for its normalising constant, it has $D+(D-1)D/2$ degrees of freedom in its parameters and it is not closed under marginalisation. 

We will now consider multi-dimensional extensions of the GvM distribution.
\section{The multivariate Generalised von Mises\label{sec:mgvm}}
In this section, we present the multivariate Generalised von Mises (mGvM) distribution as an \emph{intrinsic} circular distribution on the hyper-torus and relate it to existing distributions in the literature. Following the construction of \emph{intrinsic} distributions, the multivariate Generalised von Mises arises by constraining a $2D$-dimensional multivariate Gaussian with arbitrary mean and covariance matrix to the $D$-dimensional torus. This procedure yields the distribution
\begin{multline}
  \mgvm(\varv; \vcmean, \vcstd, \matW) \propto
    \exp
      \Big\{
      	\vcstd^{\top}\cos(\varv - \vcmean)
        \\
        - \frac{1}{2}
        \begin{bmatrix}
          \cos(\varv)\\
          \sin(\varv)
        \end{bmatrix}^{\top}
        \begin{bmatrix}
          \matW^{cc} & \matW^{cs} \\
          (\matW^{cs})^{\top} & \matW^{ss}
        \end{bmatrix}
        \begin{bmatrix}
          \cos(\varv) \\
          \sin(\varv)
        \end{bmatrix}
      \Big\}
      \label{eq:mgvm_1}
\end{multline}
where $\matW^{cc}, \matW^{cs}, \matW^{ss}$ are the blocks of the underlying Gaussian precision matrix $\matW=\cov^{-1}$, $\vcmean$ is a $D$-dimensional angle vector and $\vcstd$ is a $D$-dimensional concentration vector. \cref{eq:mgvm_1} is over-parametrised with $2D + 3(D-1)D/2$ parameters, $D$ more than the degrees of freedom of the most succinct form of the mGvM given in Supplemental Material.

The mGvM distribution generalises the multivariate von Mises by~\citet{mardia_multivariate_2008}; it collapses to the mvM when $\matW$ has the form of \cref{eq:sparsityMvM}. Whereas the one-dimensional conditionals of the mvM are von Mises and therefore unimodal and symmetric, those of the mGvM are generalised von Mises and therefore can be bimodal and asymmetric. The mGvM also captures a richer set of dependencies between the variables than the mvM, notice that the mvM is not the most general form of mGvM that has vM conditionals. The tractability of the one-dimensional conditionals of the mGvM can be leveraged for approximate inference using variational mean-field approximations and Gibbs sampling (see \cref{sec:inference}).  The mGvM is a member of the exponential family and a maximum entropy distribution subject to multidimensional first and second order circular moments constraints. We will now show that the mGvM can be used to build rich probabilistic models for circular data.
\section{Some applications of the mGvM\label{sec:applications}}
In this section, we outline two novel and important probabilistic models in which inference produces a posterior distribution that is a mGvM. The first model is a circular analogue of Gaussian Process regression and the second is a version of Principal Component Analysis for circular latent variables.
\subsection{Regression of circular data\label{sec:regression}}
Consider a regression problem in which a set of noisy output circular variables $\{\psi_n \}_{n=1}^N$ have been collected at a number of input locations $\{\vt{s}_n \}_{n=1}^N$. The treatment will apply to inputs that can be multi-dimensional and lie in any space (e.g.~they could be circular themselves). The goal is to predict circular variables $\{ \psi^{*}_m \}_{m=1}^M$ at unseen input points $\{ \vt{s}^{*}_m \}_{m=1}^M$. Here we leverage the connection between the mGvM distribution and the multivariate Gaussian in order to produce a powerful class of probabilistic models for this purpose based upon Gaussian Processes. In what follows the outputs and inputs will be represented as vectors and matrices respectively, that is $\vt{\psi}$, $\mathcal{S}$, $\vt{\psi}^*$ and $\mathcal{S}^*$.

In standard Gaussian Process regression~\cite{rasmussen_gaussian_2006} a multivariate Gaussian prior is placed over the underlying unknown function values at the input points $p(\vt{f}|\mathcal{S}) = \gp(\vt{f}; 0, \matK(\vt{s},\vt{s}^{\prime}))$, and a Gaussian noise model is assumed to produce the observations at each input location, $p(y_n|f_n,\vt{s}_n) = \mathcal{N}(y_n;f_n,\sigma_y^2)$. The prior over the function values is specified using the Gaussian Process's covariance function $K(\vt{s},\vt{s}^{\prime})$ that encapsulates prior assumptions about the properties of the underlying function, such as smoothness, periodicity, stationarity etc. Prediction then involves forming the posterior predictive distribution, $p(\vt{f}^* |\vt{y}, \mathcal{S}, \mathcal{S}^{*})$, which also takes a Gaussian form due to conjugacy.

Here an analogous approach is taken. The circular underlying function values and observations are denoted
$\varv$ and $\vt{\psi}$. The prior over the underlying function is given by a mGvM in overparametrised form
$p(\varv|\mathcal{S}) = \mgvm(\varv; 0, 0, \matK(\vt{s}, \vt{s}^{\prime})^{-1})$
and the observations are assumed to be von Mises noise corrupted versions of this function
$p(\psi_n| \arv_n, \vt{s}_n) = \vm(\psi_{n}; \arv_{n}, \cstd)$.
In order to construct a sensible prior over circular function values we use a construction that is inspired by a multi-output GP to produce bivariate variables at each input location. We then leverage the intrinsic construction of the mGvM to constrain each regressed point to the unit circle to allow the mGvM to inherit the properties from the GP covariance function it was built from. This is central to creating a flexible and powerful mGvM regression framework, as GP covariance functions that can handle exotic input variables such as circular variables, strings or graphs~\cite{gartner2003graph,duvenaud2011additive}.

Inference proceeds subtly differently to that in a GP due to an important difference between multivariate Gaussian and multivariate Generalised von Mises distributions. That is, the former are consistent under marginalisation whilst the latter are not: if a subset of mGvM variables are marginalised out, the remaining variables are not distributed according to a mGvM. Technically, this means that for analytic tractability of inference we have to handle the joint posterior predictive distribution $p(\varv, \varv^* |\vt{\psi}, \mathcal{S}, \mathcal{S}^{*})$, which is a mGvM due to conjugacy, rather than $p(\varv^* |\vt{\psi}, \mathcal{S}, \mathcal{S}^{*})$, which is not. Whilst this is somewhat less elegant than GP regression as it requires the prediction locations to be known up front, in many applications this is not a great restriction. This model type is termed transductive~\cite{quinonero2005}.
\subsection{Latent angles: dimensionality reduction and representation learning\label{sec:latentAngles}}
Next consider the task of learning the motion of an articulated rigid body from noisy measurements on a Euclidean space. Articulated rigid bodies can represent a large class of physical problems including mechanical systems, human motion and molecular interactions. The dynamics of rigid bodies can also be fully described by rotations around a fixed point plus a translation and, therefore, can be succinctly represented using angles see~\cite{chirikjian2000}. For simplicity, we will restrict our treatment to a rigid body with $D$ articulations on a 2-dimensional Euclidean space and rotations only, as the discussion trivially generalises to higher dimensional spaces and translations can be incorporated through an extra linear term. Extensions for 3-dimensional models follow directly from the 2-dimensional case, which can be seen as a first step towards these more complex models.

The Euclidean components of any point on an articulated rigid body can be described using the angles between each articulation and their distances. More precisely, for an upright, counter-clockwise coordinate system,
the horizontal and vertical components of a point in the $d$-th articulator can be written as $x_{d} = \sum_{j=1}^{d} l_{j} \sin(\vt{\varphi}_{j})$ and  $y_{d} = -\sum_{j=1}^{d} l_{j} \cos(\vt{\varphi}_{j})$, where $l_{j}$ is the length of a link $j$ to the next link or the marker. Without loss of generality, we can model only the variation around the mean angle for each joint, i.e. $\varphi_d=\arv_d-\cmean_d$ which results in the general model for noisy measurements
\begin{equation}
    \begin{bmatrix}
	  \vyy \\
      \vxx
    \end{bmatrix}
    =
    \begin{bmatrix}
      -\mat{L} \\
      \mat{L}
    \end{bmatrix}
    \begin{bmatrix}
      \cos(\vt{\varphi}) \\
      \sin(\vt{\varphi})
    \end{bmatrix}
    + \verror
    =
    \begin{bmatrix}
      \matA \\
      \matB
    \end{bmatrix}
    \begin{bmatrix}
      \cos(\varv) \\
      \sin(\varv)
    \end{bmatrix}
    + \verror
    \label{eq:rigid_body}
\end{equation}
where $\mat{L}$ is the matrix that encodes the distances between joints, $\matA$ and $\matB$ are the distance matrix rotated by the vector $\vcmean$ and $\verror\sim\gaussian(0,\stdev^2I)$.
The prior over the joint angles can be modelled by a multivariate Generalised von Mises. Here we take inspiration from Principal Component Analysis, and use independent von Mises distributions 
\begin{align}
	p(\varv_{1,\ldots,N}) &= \textstyle\prod\limits_{n = 1}^{N}\prod\limits_{d = 1}^{D}
    	\vm\left(\arv_{d,n};0,\cstd_{d}\right).
        \label{eq:vonMisesPriors}
\end{align}
Due to conjugacy, the posterior distribution over the latent angles is a mGvM distribution. This can be informally verified by noting that the priors on the latent angles $\varv$ are exponentials of linear functions of sines and cosines, while the likelihood is the exponential of a quadratic function in sine and cosines. This leads to the posterior being an exponential quadratic function of sines and cosines and, hence, mGvM.

The model can be extended to treat the parameters in Bayesian way by including sparse priors over the coefficient matrices $\matA$ and $\matB$ and the observation noise. A standard choice for this task is to define Automatic Relevance Detection priors~\cite{mackay_bayesian_1994} over the columns of these matrices defined
as $\gaussian(\matA_{m,d};0,\stdev^{2}_{\matA,d})$ and $\gaussian(\matB_{m,d};0,\stdev^{2}_{\matB,d})$ in order to perform automatic structure learning. Additional Inverse Gamma priors over $\stdev^{2}_{\matA,d}$, $\stdev^{2}_{\matB,d}$ and $\stdev^{2}$ are also employed.

The dimensionality of the latent angle space can be lower than the dimensionality of the observed space, in which case learning and inference perform dimensionality reduction that maps real-valued data to a lower-dimensional torus. Besides motion capture, toroidal manifolds can also prove useful when modelling other relevant applications, such as electroencephalogram (EEG) and audio signals~\cite{turner_probabilistic_2011}. Further connections between dimensionality reduction with the mGvM and Probabilistic Principal Component Analysis (PPCA) proposed by~\citet{tipping_probabilistic_1999} (including limiting behaviour and geometrical relations between these models) are explored in the Supplementary Material. As a consequence of these similarities, we denote this model as circular Principal Component Analysis (cPCA).
\section{Approximate inference for the mGvM\label{sec:inference}}
The multivariate Generalised von Mises does not admit an analytic expression for its normalizing constant, therefore we need to resort to approximate inference techniques. This section presents two approaches that exploit the tractable univariate conditionals of the mGvM: Gibbs sampling and mean-field variational inference.
\subsection{Gibbs sampling}
A Gibbs sampling procedure for sampling the mGvM of Equation \eqref{eq:mgvm_1} can be derived leveraging the GvM form of the one-dimensional mGvM conditionals. In particular, the Gibbs sampler updates for the $d$-th conditional of the mGvM will have the form
\begin{equation}
  p(\arv_{d}|\varv_{\neq d}) = \gvm(\arv_{d}; \tilde{\cstd}_{1,d}, \cstd_{2,d},\tilde{\cmean}_{1,d}, \cmean_{2,d})
\end{equation}
where $\tilde{\cstd}_{1,d}$ and $\tilde{\cmean}_{1,d}$ are functions of $\vcstd$, $\vcmean$ and $\varv_{\neq d}$ given in the Supplementary Material.

The Gibbs sampler can be used to support approximate maximum-likelihood learning by using it to compute the expectations required by the EM algorithm~\cite{wei_tanner_1990}. However, it is well-known that Gibbs sampling becomes less effective as the joint distribution becomes more correlated and the dimensionality grows. This is particularly significant when using the distribution in high-dimensional cases with rich correlational structure, such as those considered later in the paper.
\subsection{Mean-field Variational inference}
As a consequence of the problems encountered when using Gibbs sampling, the variational inference framework emerges as an attractive, scalable approach to handling inference in when the posterior distribution is a mGvM.

The variational inference framework~\cite{jordan1999} aims to approximate an intractable posterior $p(\varv|\vt{\psi}, \theta)$ with a distribution $q(\varv|\rho)$ by minimising the Kullback-Leiber divergence from the distribution $q$ to $p$. If the approximating distribution is chosen to be fully factored, i.e. $\q(\varv) = \prod_{d=1}^{d}\q_{d}(\arv_{d})$, the optimal functional form for $\q_{d}(\arv_{d})$ can be obtained analytically using calculus of variations. The functional form of each mean-field factor is inherited from the one-dimensional conditionals and consequently is a Generalised von Mises of the form 
\begin{equation*}
\q_{d}(\arv_{d}) = \gvm(\arv_{d}; \bar{\cstd}_{1,d}, \cstd_{2,d},\bar{\cmean}_{1,d}, \cmean_{2,d})
\end{equation*}
where the formulas for the parameters $\bar{\cstd}_{1,d}$ and $\bar{\cmean}_{1,d}$ are similar in nature to the Gibbs sampling update and given in the Supplementary Material.

Furthermore, since the moments of the Generalised von Mises can be computed through series approximations~\cite{gatto_computational_2008}, the errors from series truncation are negligible if a sufficiently large number of terms is considered. It is possible to obtain gradients of the variational free energy and optimise it with standard optimisation methods such as Conjugate Gradient or Quasi-Newton methods instead of resorting to coordinate-ascent under the variational Expectation-Maximization algorithm which often is slow to converge.

Despite these improvements, we found empirically that accurate calculations of the moments of a Generalised von Mises distribution can become costly when the magnitude of the concentration parameters exceeds $\approx$100 and the posterior concentrates. This numerical instability occurs when the infinite expansion for computing the moments contains a large number of significant terms that have alternating signs leading to accumulation of numerical errors. It is possible to use other approximate integrations schemes if these cases arise during inference. An alternative way to alleviate this problem is to consider a sub-optimal form of factorised approximating distribution. An obvious choice is to use von Mises factors as this results in tractable updates and requires simpler moment calculations. A von Mises field can also be motivated as a first order approximation to a GvM field by requiring that the log approximating distribution is linear in sine and cosine terms, as shown in the Supplementary Material.

In addition to inference, we can use the same variational framework for learning in cases where the mGvM we wish to approximate is a posterior of tractable likelihoods and priors, as in the cPCA model. To achieve this, we form the variational free-energy lower bound on the log-marginal likelihood as
\begin{equation*}
  \log p(\vt{\psi}|\theta) \geq \free(q,\theta) = \expect{\log p(\vt{\psi},\varv| \theta)}_{q(\varv|\rho)} + \entropy(q),
\end{equation*}
where $\entropy(q)$ is the entropy of the approximating distribution and $q(\varv|\rho)$, $p(\varv, \vt{\psi}| \theta)$ is the model log-joint distribution, $\free(q,\theta,\rho)$ is the variational free-energy, $\theta$ are the model parameters and $\rho$ represents the parameters of the approximating distribution. The same bound cannot be used directly for doubly-intractable mGvM models, such as the circular regression model, and it constitutes an area for further work.
\section{Experimental results\label{sec:results}}
To demonstrate approximate inference on the applications outlined in \cref{sec:applications} we present experiments on synthetic and real datasets. A comprehensive description and the data sets used in the all experiments conducted are available at \url{http://tinyurl.com/mgvm-release}. Further experimental details are also provided in the Supplementary Material.
\subsection{Comparison to other circular distributions}
\begin{figure*}[tb]
  \centering
  \begin{tabular}{ccccc}
    \raisebox{-0.5\height}{\includegraphics[width=0.18\textwidth,clip=true,trim=3mm 3mm 3mm 3mm]{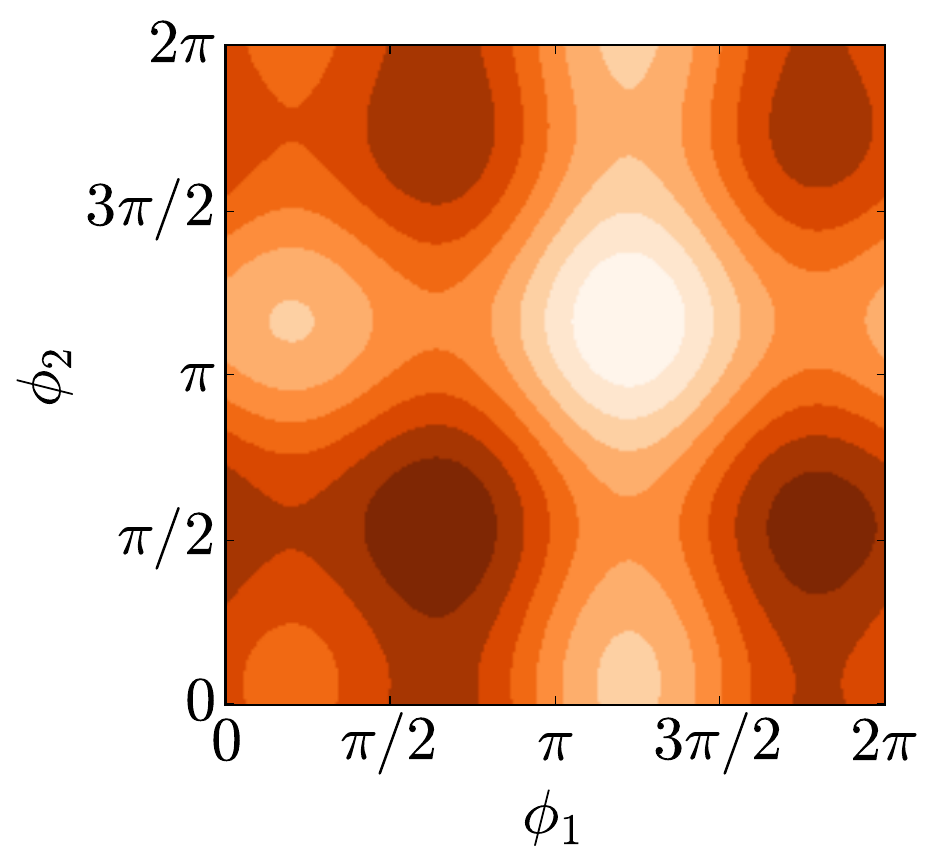}} &
    \raisebox{-0.5\height}{\includegraphics[width=0.18\textwidth,clip=true,trim=3mm 3mm 3mm 3mm]{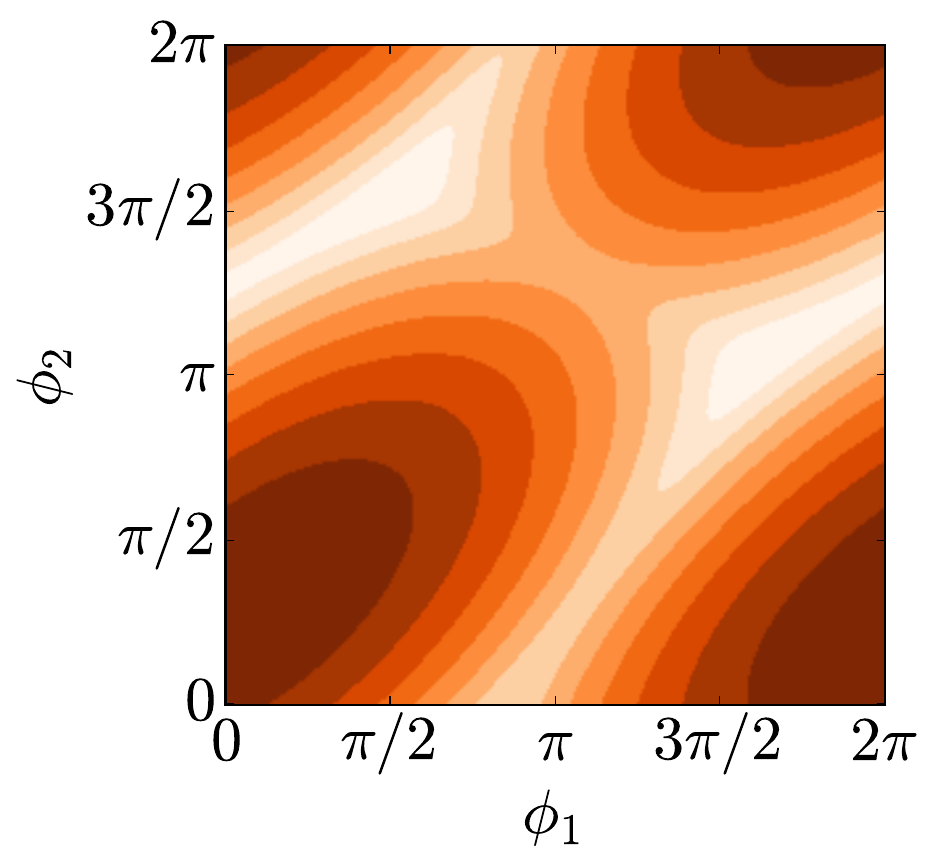}} &
    \raisebox{-0.5\height}{\includegraphics[width=0.18\textwidth,clip=true,trim=3mm 3mm 3mm 3mm]{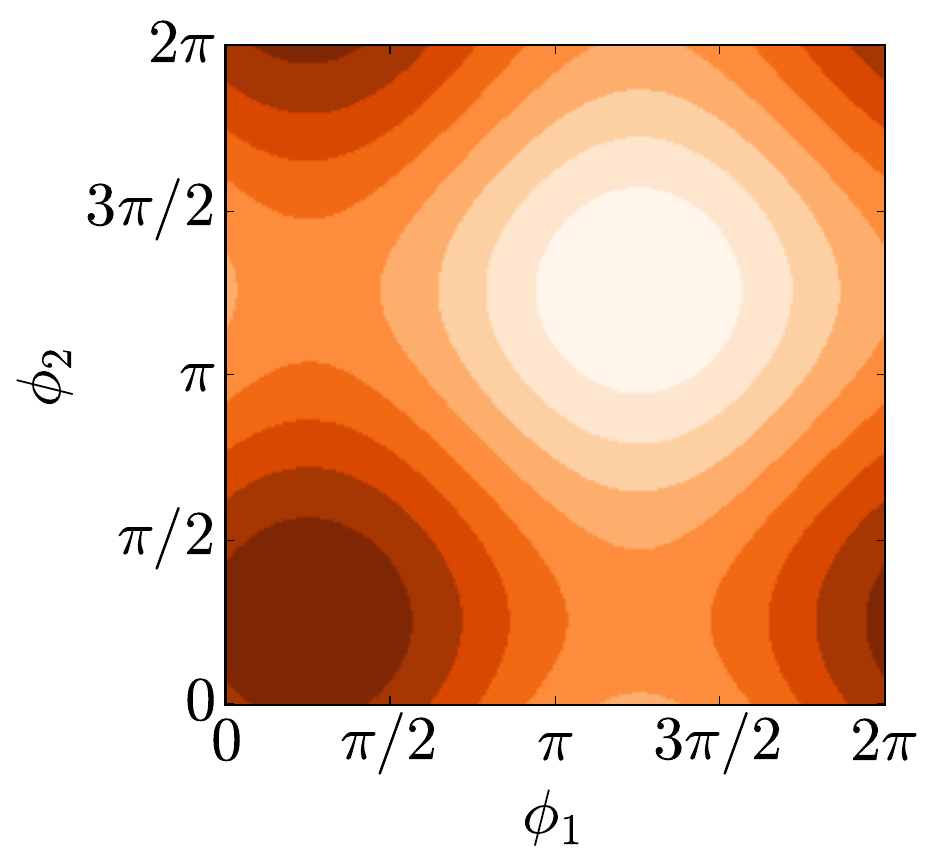}} &
    \raisebox{-0.5\height}{\includegraphics[width=0.18\textwidth,clip=true,trim=3mm 3mm 3mm 3mm]{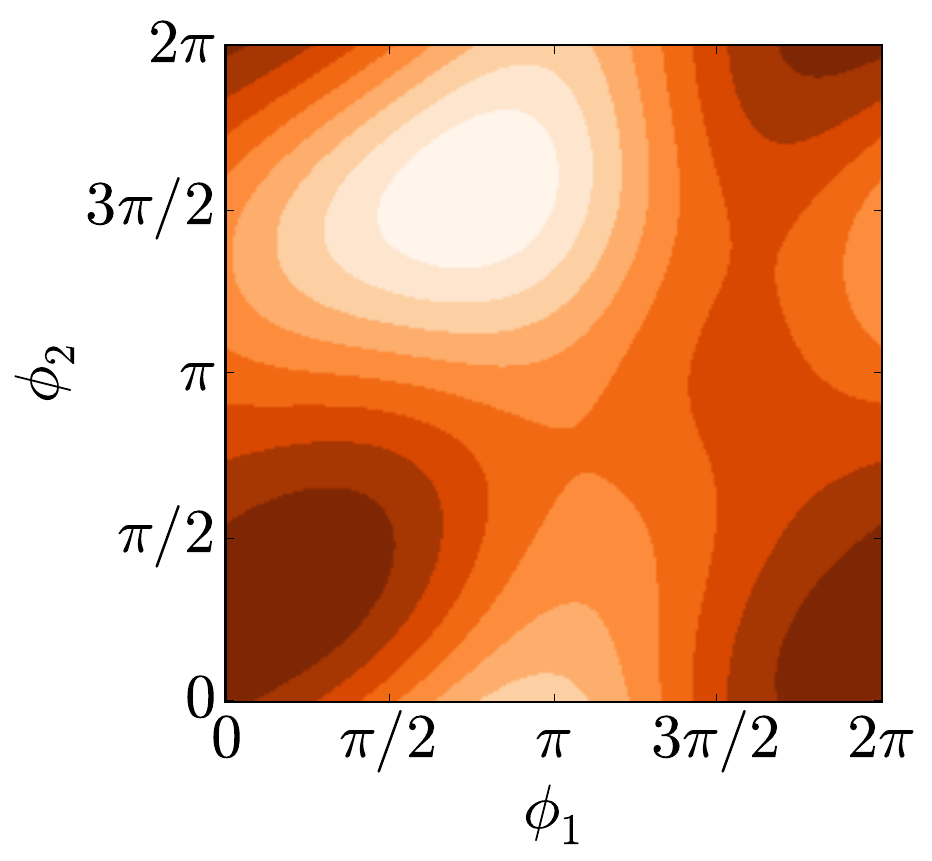}} &
    \raisebox{-0.5\height}{\includegraphics[width=0.18\textwidth,clip=true,trim=3mm 3mm 3mm 3mm]{mvm-approx.pdf}} \\
    \hspace{6mm}\small{(a)} & \hspace{6mm}\small{(b)} & \hspace{6mm}\small{(c)} & \hspace{6mm}\small{(d)} & \hspace{6mm}\small{(e)}
  \end{tabular}
    \caption{Circular distribution approximations: a base mGvM (a) and its optimal approximations using a multivariate wrapped Gaussian (b) and the mvM (c). The mGvM approximation to the mWG (b) is presented in (d) and the mGvM approximation to the mvM in (c) is presented in (e). Neither the multivariate wrapped Gaussian nor the mvM can accommodate for the asymmetries and the multiple modes of the mGvM, however, the mGvM is able to approximate the mWG high-probability regions and fully recover the mvM. Darker regions have higher probability than lighter regions.}
    \label{fig:mgvm-comparison}
\end{figure*}

For illustrative purposes we qualitatively compared multivariate Wrapped Gaussian and mvM approximations to a base mGvM and mGvM approximations to these two distributions. The approximations were obtained by numerically minimising the KL divergence between the approximating distribution and the base distributions. These experiments were conducted on a two-dimensional setting in order to render the computation of the normalising constant of the mGvM and the mvM tractable by numerical integration. The resulting distributions are shown in \cref{fig:mgvm-comparison}.

In \cref{fig:mgvm-comparison}, the mvM and the multivariate wrapped Gaussian cannot capture the multimodality and asymmetry of the mGvM. Moreover, these distributions approximate the multiple modes by increasing their variance and assigning high probability to the region of low-probability between the modes of the mGvM. On the other hand, when the mvM and the multivariate wrapped Gaussian are approximated by the mGvM, the mGvM is able to approximate well the high-probability zones of the wrapped Gaussian and its unimodality and fully recover the mvM.

We also compared the performance of Gibbs sampling and variational inference for a bivariate GvM. To compare the approximate inference procedures, we analysed the run time for each method and the error it produced in terms of the KL divergence between the true distribution and the approximations on a discretized grid. The Gibbs sampling procedure required a total of 3466 samples and \SI{3.1}{\s} to achieve the same level of error as the variational approximation achieved after \SI{0.02}{\s}. The variational approach was considerably more efficient than Gibbs sampling, and theory suggests this behaviour holds for higher dimensions, see~\citet{david_mackay_information_2003}.
\subsection{Regression with the mGvM}
In this section, we investigate the advantages of employing the mGvM regression model discussed in \cref{sec:regression} over two common approaches to handling circular data in machine learning contexts.

The first approach is to ignore the circular nature of the data and fit a non-circular model. This approach is not infrequent as it is reasonable in contexts where angles are constrained to a subset of the unit circle and there is no wrappping. A typical example of the motivation for such models is the use of a first-order Taylor approximation to the rate of change of an angle as can be found in classical aircraft control applications. To represent this approach to modeling, we will fit a one-dimensional GP (1D-GP) to the data sets.

The second approach tries to address the circular behaviour by regressing the sine and cosine of the data. In this approach, the angle can be extracted by taking the arc tangent of the ratio between sine and cosine components. While this approach partially addresses the underlying topology of the data, the uncertainty estimates for a non-circular model can be poorly calibrated. Here, each data point is modeled by a two-dimensional vector with the sine and cosine of each data point using a two-dimensional GP (2D-GP).

Five data sets were used in this evaluation. A toy data set generated by wrapping a Mexican hat function around the unit circle, a dataset consisting Uber ride requests in NYC in April 2014\footnote{\url{https://github.com/fivethirtyeight/uber-tlc-foil-response}}, the tide levels predictions from the UK Hydrographic Office in 2016\footnote{\url{http://www.ukho.gov.uk/Easytide/easytide/SelectPort.aspx}} as function of the latitude and longitude of a given port, the first side chain angle of aspartate as a function of backbone angles in proteins~\cite{harder_beyond_2010}, and yeast cell cycle phase as a function of gene expression~\cite{santos2015}.

To assess how well the fitted models approximate the distribution of the data, a subset of the data points was kept for validation and the models scored in terms of the log likelihood of the validation data set. To guarantee fairness in the comparison, the likelihood of the 2D-GP was projected back to the unit circle by marginalising the radial component of the model for each point. This converts the 2D-GP into a one-dimensional projected Gaussian distribution over angles. The results are summarised in \cref{tab:regression}.

\makeatletter
\newcolumntype{B}[3]{>{\boldmath\DC@{#1}{#2}{#3}}c<{\DC@end}}
\makeatother
\newcolumntype{d}{D{.}{.}{2.7}}
\newcolumntype{b}{B{.}{.}{2.7}}
\newcommand{\bcell}[1]{\multicolumn{1}{b}{#1}}
\begin{table}[tb]
  \caption{Log-likelihood score for regression with the mGvM, 1D-GP and 2D-GP on validation data.}
  \label{tab:regression}
  \centering
  \begin{tabular}{lddd}
    \toprule
    Data set                   & \multicolumn{1}{c}{mGvM} & \multicolumn{1}{c}{1D-GP} & \multicolumn{1}{c}{2D-GP} \\
    \midrule
    Toy                        &       \bcell{2.02\E{4}}  &                -1.62\E{3} &        8.28\E{2} \\
    Uber                       &       \bcell{3.29\E{4}}  &                -1.49\E{3} &       -2.83\E{2} \\
    Tides                      &       \bcell{1.25\E{4}}  &                -6.46\E{4} &       -8.41\E{1} \\
    Protein                    &       \bcell{1.42\E{5}}  &                -3.34\E{5} &        1.28\E{5} \\
    Yeast                      &       \bcell{1.33\E{2}}  &                -1.46\E{2} &       -1.65\E{1} \\
    \bottomrule
  \end{tabular}
\end{table}

The results shown in \cref{tab:regression} indicate that the mGvM provides a better overall fit than the 1D-GP and the 2D-GP in all experiments. The 1D-GP approach performs poorly in every case studied as it cannot account for the wrapping behaviour of circular data. The 2D-GP performs better than the 1D-GP, however in the Uber, Tides and Yeast datasets its performance is substantially closer to the one presented by the 1D-GP case rather than the mGvM. The toy dataset is examined in \cref{fig:1d_regression}, showing the 2D-GP learns a different underlying function and cannot capture bimodality.

\begin{figure}[t]
  \includegraphics[width=0.45\columnwidth,clip=true,trim=5mm 6mm 3mm 5mm]{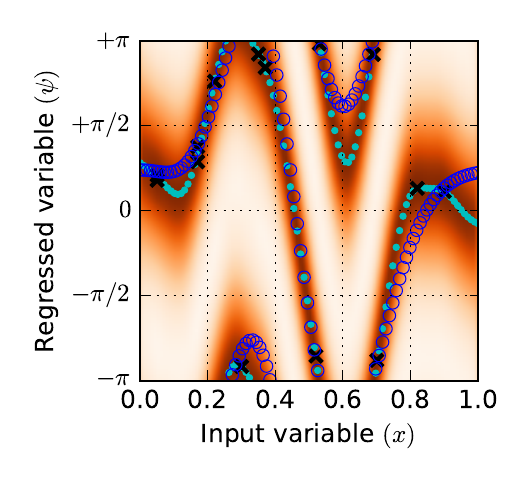}\includegraphics[width=0.45\columnwidth,clip=true,trim=5mm 6mm 3mm 5mm]{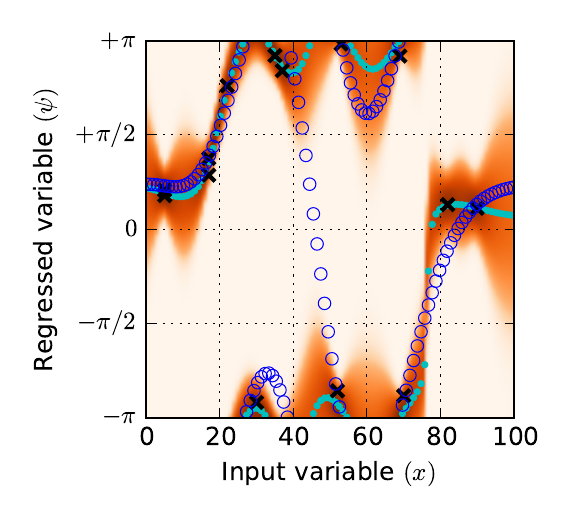}
  \caption{Regression on a toy data set using the mGvM (left) and 2D GP (right): data points are denoted by crosses, the true function by circles and predictions by solid dots.}
  \label{fig:1d_regression}
\end{figure}

\subsection{Dimensionality reduction}
To demonstrate the dimensionality reduction application, we analysed two datasets: one motion capture dataset comprising marker positions placed on a subject's arm and captured through a low resolution camera and another set comprising of a noisy simulation of a 4-DOF robot arm under the same motion capture conditions.

We compared the model using point estimates for the matrices $\matA$ and $\matB$, a variational Bayes approach by including ARD priors for $\matA$ and $\matB$, Probabilistic Principal Component Analysis (PPCA)~\cite{tipping_probabilistic_1999} and the Gaussian Process Latent Variable Model (GP-LVM)~\cite{lawrence2004gpml} using a squared exponential kernel and a linear kernel.
The models using the mGvM require special attention to initialisation. To initialise the test, we used a greedy clustering algorithm to estimate the matrices $\matA$ and $\matB$. The variational Bayes model was initialised using the learned parameters for the point estimate model.

The performance of each model was assessed by denoising the original dataset corrupted by additional Gaussian noise of 2.5, 5 and 10 pixels and comparing the signal-to-noise ratio (SNR) on a test dataset. The best results after initializing the models at 3 different initial starting points are summarized in \cref{tab:mocap-results} and additional experiments for a wider range of noise levels are available in the Supplemental Material.
In \cref{tab:mocap-results}, the point estimate cPCA model performs best and is followed by its variational Bayes version for both datasets (the poor performance of the variational Bayes version is likely to be due to biases that can affect variational methods \cite{turner-and-sahani:2011a}). In the motion capture dataset, the latent angles are highly concentrated. Under these circumstances, the small-angle approximation for sine and cosine provides good results and the cPCA model degenerates into the PPCA model as shown in the Supplementary Material. This behaviour is reflected in the proximity of the PPCA and cPCA signal to noise ratios in \cref{tab:mocap-results}. In the robot dataset, the latent angles are less concentrated. As a result, the behaviour of the PPCA and cPCA models is different which explains the larger gap between the results obtained for these models.

\makeatletter
\newcolumntype{B}[3]{>{\boldmath\DC@{#1}{#2}{#3}}c<{\DC@end}}
\makeatother
\newcolumntype{d}{D{.}{.}{1.8}}
\newcolumntype{b}{B{.}{.}{1.8}}
\begin{table}[tb]
  \caption{Signal-to-noise ratio (dB) of the learned latent structure after denoising corrupted signals with by Gaussian noise.}
  \label{tab:mocap-results}
  \centering
  \begin{tabular}{lcccccc}
    \toprule
    \multirow{2}{*}{Model}      & \multicolumn{3}{c}{Motion Capture} & \multicolumn{3}{c}{Robot}\\
    \cmidrule(lr){2-4}\cmidrule(lr){5-7}
                                & \multicolumn{1}{c}{2.5} & \multicolumn{1}{c}{5} & \multicolumn{1}{c}{10} & \multicolumn{1}{c}{2.5} & \multicolumn{1}{c}{5} & \multicolumn{1}{c}{10} \\
    \midrule
    cPCA-Point                & \textbf{29.6} & \textbf{23.5} & \textbf{17.6} & \textbf{33.5} & \textbf{30.0} & \textbf{24.9} \\
    cPCA-VB                   &        24.6   &         21.9  &         17.6  &         33.2  &         29.8  &         24.8  \\
    PPCA                      &        23.6   &         20.9  &         17.2  &         22.3  &         21.8  &         20.5  \\
    GPLVM-SE                  &         8.6   &          8.5  &          8.2  &         21.8  &         15.7  &         15.2  \\
    GPLVM-L                   &        11.0   &          7.5  &          8.1  &         24.0  &         16.6  &         15.9  \\
    \bottomrule
  \end{tabular}
\end{table}
\section{Conclusions}
In this paper we have introduced the multivariate Generalised von Mises, a new circular distribution with novel applications in circular regression and circular latent variable modelling in a first attempt to close the gap between circular statistics and the machine learning communities. We provided a brief review of the construction of circular distributions including the connections between the Gaussian distribution and the multivariate Generalised von Mises. We provided a scalable way to perform inference on the mGvM model through the variational free energy framework and demonstrated the advantages of the mGvM over GP and mvM through a series of experiments.

\section{Acknowledgements}
AKWN thanks CAPES grant BEX 9407-11-1. JF thanks the Danish Council for Independent Research grant 0602-02909B. RET thanks EPSRC grants EP/L000776/1 and EP/M026957/1.

\bibliography{references}

\begin{thebibliography}{}

\bibitem[\protect\citeauthoryear{Ben-Yishai, Bar-Or, and
  Sompolinsky}{1995}]{benyishai1995}
Ben-Yishai, R.; Bar-Or, R.; and Sompolinsky, H.
\newblock 1995.
\newblock Theory of orientation tuning in visual cortex.
\newblock {\em Proceedings of the National Academy of Sciences of the United
  States of America} 92(9):3844—3848.

\bibitem[\protect\citeauthoryear{Boomsma \bgroup et al\mbox.\egroup
  }{2008}]{boomsma08}
Boomsma, W.; Mardia, K.~V.; Taylor, C.~C.; {Ferkinghoff-Borg}, J.; Krogh, A.;
  and Hamelryck, T.
\newblock 2008.
\newblock A generative, probabilistic model of local protein structure.
\newblock {\em Proceedings of the National Academy of Sciences}
  105(26):8932--8937.

\bibitem[\protect\citeauthoryear{Brunsdon and
  Corcoran}{2006}]{brunsdon_using_2006}
Brunsdon, C., and Corcoran, J.
\newblock 2006.
\newblock Using circular statistics to analyse time patterns in crime
  incidence.
\newblock {\em Computers, Environment and Urban Systems} 30(3):300--319.

\bibitem[\protect\citeauthoryear{Chirikjian and Kyatkin}{2000}]{chirikjian2000}
Chirikjian, G.~S., and Kyatkin, A.~B.
\newblock 2000.
\newblock {\em {Engineering Applications of Noncommutative Harmonic Analysis:
  With Emphasis on Rotation and Motion Groups}}.
\newblock Abingdon: CRC Press.

\bibitem[\protect\citeauthoryear{{David
  MacKay}}{2003}]{david_mackay_information_2003}
{David MacKay}.
\newblock 2003.
\newblock {\em Information Theory, Inference, and Learning Algorithms}.
\newblock Cambridge: Cambridge University Press, 2 edition.

\bibitem[\protect\citeauthoryear{Duvenaud, Nickisch, and
  Rasmussen}{2011}]{duvenaud2011additive}
Duvenaud, D.~K.; Nickisch, H.; and Rasmussen, C.~E.
\newblock 2011.
\newblock Additive gaussian processes.
\newblock In Shawe-Taylor, J.; Zemel, R.~S.; Bartlett, P.~L.; Pereira, F.; and
  Weinberger, K.~Q., eds., {\em Advances in Neural Information Processing
  Systems 24}. Curran Associates, Inc.
\newblock  226--234.

\bibitem[\protect\citeauthoryear{Ferrari}{2009}]{ferrari_wrapping_2009}
Ferrari, C.
\newblock 2009.
\newblock {\em The Wrapping Approach for Circular Data Bayesian Modelling}.
\newblock Ph.D. Dissertation, Universit\`{a} di Bologna, Bologna.

\bibitem[\protect\citeauthoryear{Frellsen \bgroup et al\mbox.\egroup
  }{2009}]{frellsen09}
Frellsen, J.; Moltke, I.; Thiim, M.; Mardia, K.~V.; {Ferkinghoff-Borg}, J.; and
  Hamelryck, T.
\newblock 2009.
\newblock A probabilistic model of {RNA} conformational space.
\newblock {\em {PLoS} Computational Biology} 5(6):e1000406.

\bibitem[\protect\citeauthoryear{Gao \bgroup et al\mbox.\egroup
  }{2010}]{gao2010}
Gao, S.; Hartman, John~L, I.; Carter, J.~L.; Hessner, M.~J.; and Wang, X.
\newblock 2010.
\newblock Global analysis of phase locking in gene expression during cell
  cycle: the potential in network modeling.
\newblock {\em BMC Systems Biology} 4(1):167.

\bibitem[\protect\citeauthoryear{G{\"a}rtner, Flach, and
  Wrobel}{2003}]{gartner2003graph}
G{\"a}rtner, T.; Flach, P.; and Wrobel, S.
\newblock 2003.
\newblock On graph kernels: Hardness results and efficient alternatives.
\newblock In {\em Learning Theory and Kernel Machines}. Springer Berlin
  Heidelberg.
\newblock  129--143.

\bibitem[\protect\citeauthoryear{Gatto and
  Jammalamadaka}{2007}]{gatto_generalized_2007}
Gatto, R., and Jammalamadaka, S.~R.
\newblock 2007.
\newblock The generalized von mises distribution.
\newblock {\em Statistical Methodology} 4(3):341--353.

\bibitem[\protect\citeauthoryear{Gatto}{2008}]{gatto_computational_2008}
Gatto, R.
\newblock 2008.
\newblock Some computational aspects of the generalized von mises distribution.
\newblock {\em Statistics and Computing} 18(3):321--331.

\bibitem[\protect\citeauthoryear{Harder \bgroup et al\mbox.\egroup
  }{2010}]{harder_beyond_2010}
Harder, T.; Boomsma, W.; Paluszewski, M.; Frellsen, J.; Johansson, K.~E.; and
  Hamelryck, T.
\newblock 2010.
\newblock Beyond rotamers: a generative, probabilistic model of side chains in
  proteins.
\newblock {\em {BMC} Bioinformatics} 11(1):306.

\bibitem[\protect\citeauthoryear{Jona-Lasinio, Gelfand, and
  Jona-Lasinio}{2012}]{jona-lasinio2012}
Jona-Lasinio, G.; Gelfand, A.; and Jona-Lasinio, M.
\newblock 2012.
\newblock Spatial analysis of wave direction data using wrapped gaussian
  processes.
\newblock {\em The Annals of Applied Statistics} 6(4):1478--1498.

\bibitem[\protect\citeauthoryear{Jordan \bgroup et al\mbox.\egroup
  }{1999}]{jordan1999}
Jordan, M.~I.; Ghahramani, Z.; Jaakkola, T.~S.; and Saul, L.~K.
\newblock 1999.
\newblock An introduction to variational methods for graphical models.
\newblock {\em Machine Learning} 37(2):183--233.

\bibitem[\protect\citeauthoryear{Lawrence}{2004}]{lawrence2004gpml}
Lawrence, N.~D.
\newblock 2004.
\newblock Gaussian process latent variable models for visualisation of high
  dimensional data.
\newblock In Thrun, S.; Saul, L.; and Sch\"{o}lkopf, B., eds., {\em Advances in
  Neural Information Processing Systems 16}. {MIT} Press.
\newblock  329--336.

\bibitem[\protect\citeauthoryear{Lebanon}{2005}]{lebanon2005}
Lebanon, G.
\newblock 2005.
\newblock {\em Riemannian Geometry and Statistical Machine Learning}.
\newblock Ph.D. Dissertation, Carnegie Mellon University, Pittsburg.

\bibitem[\protect\citeauthoryear{MacKay}{1994}]{mackay_bayesian_1994}
MacKay, D. J.~C.
\newblock 1994.
\newblock Bayesian non-linear modelling for the prediction competition.
\newblock In {\em {ASHRAE} Transactions, V.100, Pt.2},  1053--1062.
\newblock {ASHRAE}.

\bibitem[\protect\citeauthoryear{Mardia and
  Jupp}{2000}]{mardia_directional_2000}
Mardia, K.~V., and Jupp, P.~E.
\newblock 2000.
\newblock {\em Directional statistics}.
\newblock Chichester; New York: J. Wiley.

\bibitem[\protect\citeauthoryear{Mardia \bgroup et al\mbox.\egroup
  }{2008}]{mardia_multivariate_2008}
Mardia, K.~V.; Hughes, G.; Taylor, C.~C.; and Singh, H.
\newblock 2008.
\newblock A multivariate von mises distribution with applications to
  bioinformatics.
\newblock {\em Canadian Journal of Statistics} 36(1):99--109.

\bibitem[\protect\citeauthoryear{Qui{\~n}onero-Candela and
  Rasmussen}{2005}]{quinonero2005}
Qui{\~n}onero-Candela, J., and Rasmussen, C.~E.
\newblock 2005.
\newblock A unifying view of sparse approximate gaussian process regression.
\newblock {\em Journal of Machine Learning Research} 6:1939--1959.

\bibitem[\protect\citeauthoryear{Rasmussen and
  Williams}{2006}]{rasmussen_gaussian_2006}
Rasmussen, C.~E., and Williams, C. K.~I.
\newblock 2006.
\newblock {\em Gaussian processes for machine learning}.
\newblock {MIT} Press.

\bibitem[\protect\citeauthoryear{Santos, Wernersson, and
  Jensen}{2015}]{santos2015}
Santos, A.; Wernersson, R.; and Jensen, L.~J.
\newblock 2015.
\newblock Cyclebase 3.0: a multi-organism database on cell-cycle regulation and
  phenotypes.
\newblock {\em Nucleic Acids Research} 43(D1):D1140--D1144.

\bibitem[\protect\citeauthoryear{Tipping and
  Bishop}{1999}]{tipping_probabilistic_1999}
Tipping, M.~E., and Bishop, C.~M.
\newblock 1999.
\newblock Probabilistic principal component analysis.
\newblock {\em Journal of the Royal Statistical Society: Series B (Statistical
  Methodology)} 61(3):611--622.

\bibitem[\protect\citeauthoryear{Turner and
  Sahani}{2011a}]{turner_probabilistic_2011}
Turner, R., and Sahani, M.
\newblock 2011a.
\newblock Probabilistic amplitude and frequency demodulation.
\newblock In {\em Advances in Neural Information Processing Systems 24},
  981--989.
\newblock {MIT} Press.

\bibitem[\protect\citeauthoryear{Turner and
  Sahani}{2011b}]{turner-and-sahani:2011a}
Turner, R.~E., and Sahani, M.
\newblock 2011b.
\newblock Two problems with variational expectation maximisation for
  time-series models.
\newblock In Barber, D.; Cemgil, T.; and Chiappa, S., eds., {\em Bayesian Time
  series models}. Cambridge University Press.
\newblock chapter~5,  109--130.

\bibitem[\protect\citeauthoryear{Wadhwa \bgroup et al\mbox.\egroup
  }{2013}]{wadhwa2013}
Wadhwa, N.; Rubinstein, M.; Durand, F.; and Freeman, W.~T.
\newblock 2013.
\newblock Phase-based video motion processing.
\newblock {\em ACM Transactions on Graphics} 32(4).

\bibitem[\protect\citeauthoryear{Wang and
  Gelfand}{2013}]{wang_directional_2013}
Wang, F., and Gelfand, A.~E.
\newblock 2013.
\newblock Directional data analysis under the general projected normal
  distribution.
\newblock {\em Statistical methodology} 10(1):113--127.

\bibitem[\protect\citeauthoryear{Wei and Tanner}{1990}]{wei_tanner_1990}
Wei, G. C.~G., and Tanner, M.~A.
\newblock 1990.
\newblock A monte carlo implementation of the em algorithm and the poor man's
  data augmentation algorithms.
\newblock {\em Journal of the American Statistical Association}
  85(411):699--704.

\end{thebibliography}
\bibliographystyle{aaai}

\pagebreak
\appendix
\onecolumn
\section{Supplementary Material for \\The Multivariate Generalised\ von Mises\\ Distribution: Inference and application}

\subsection{Diagramatic view of circular distributions genesis\label{sec:circgenesis}}

\begin{figure}[htb]
  \centering
  \includegraphics{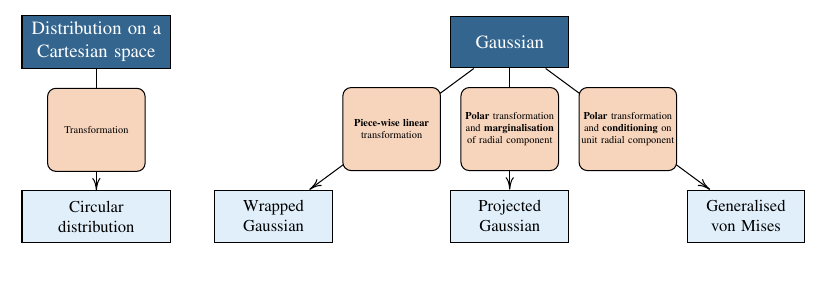}
  \caption{Graphical summary of the genesis of circular distributions through transformations of Euclidean distributions.}
  \label{fig:circDistSummary}
\end{figure}

The discussion on circular distributions genesis on the main paper can be diagramatically sumarised as \cref{fig:circDistSummary}.

\subsection{Derivation of the Multivariate Generalised von Mises\label{sec:genesis}}

The Multivariate Generalised von Mises distribution can be derived by applying the polar transformation to a $2D$-dimensional multivariate Gaussian distribution and conditioning $D$-pairs to the unit circle. Since order in which we conduct these two operations is interchangeable, we will first condition its pairs to the unit circle and then apply the polar transformation.

More precisely, we assume that $\vxx \in \reals^{2D}$, such that $x_{d}^2 + x_{D + d}^2 = 1$ for $d=1, \ldots, D$. In this case, the polar transformation of $\vxx$ allows us to write $x_{d} = \cos(\arv_{d})$, $x_{D + d} = \sin(\arv_{d})$. Furthermore, without loss of generality, the mean $\vmean$ of a multivariate Gaussian will also be constrained to the unit circle and can be parametrised in terms of angles $\vcmean$ so that $\mean_{d}=\cos(\cmean_n)$, $\mean_{D + d}=\sin(\cmean_{D + d})$.

Now let and $\matW = \cov^{-1}$ be the inverse covariance matrix of a multivariate Gaussian. Using the parametrisation of $\vxx$ and $\vmean$ in terms of $\varv$ and $\vcmean$, we can expand the quadratic in the exponential of the multivariate Gaussian into
\begin{align}
  \begin{bmatrix}
    \cos(\varv)-\cos(\vcmean)\\
    \sin(\varv)-\sin(\vcmean)
  \end{bmatrix}^{\top}
  &\matW
  \begin{bmatrix}
    \cos(\varv)-\cos(\vcmean) \\
    \sin(\varv)-\sin(\vcmean)
  \end{bmatrix}
  \label{eq:expanded_prod1}
                \\
        &= \sum_{d=1}^{D} w_{d,d}(\cos(\arv_{d})-\cos(\cmean_{d}))^2
                           + w_{D+n,D+d}(\sin(\arv_{d})-\sin(\cmean_{d}))^2
                \notag
                \\
        &+ \sum_{d=1}^{D}\sum_{\jj=1}^{d-1}
                w_{d, \jj}(\cos(\arv_{d})-\cos(\cmean_{d}))(\cos(\arv_{\jj})-\cos(\cmean_{\jj}))
                \notag
                \\
        &+ \sum_{d=1}^{D}\sum_{\jj=1}^{D}
                w_{d, D+\jj}(\cos(\arv_{d})-\cos(\cmean_{d}))(\sin(\arv_{\jj})-\sin(\cmean_{\jj}))
                \notag
                \\
        &+ \sum_{d=1}^{D}\sum_{\jj=1}^{D}
                w_{D+d,\jj}(\sin(\arv_{d})-\sin(\cmean_{d}))(\cos(\arv_{\jj})-\cos(\cmean_{\jj}))
                \notag
                \\
        &+ \sum_{d=1}^{D}\sum_{\jj=1}^{d-1}
                w_{D+d,D+\jj}(\sin(\arv_{d})-\sin(\cmean_{d}))(\sin(\arv_{\jj})-\sin(\cmean_{\jj}))
                \notag.
\end{align}

The sums on the RHS in Equation \eqref{eq:expanded_prod1} can be expanded into
\begin{align}
        &\sum_{d=1}^{D} w_{d,d}(\cos(\arv_{d})^2 - 2\cos(\arv_{d})\cos(\cmean_{d}) -\cos(\cmean_{d})^2)
                \label{eq:expanded_prod2}
                \\
        &+ \sum_{d=1}^{D}w_{D+n,D+d}(\sin(\arv_{d})^2 - 2\sin(\arv_{d})\sin(\cmean_{d}) -\sin(\cmean_{d})^2)
                \notag
                \\
        &+2\sum_{d=1}^{D}\sum_{\jj=1}^{d-1}
                w_{d, \jj}(\cos(\arv_{d})\cos(\arv_{\jj})-\cos(\cmean_{d})\cos(\arv_{\jj})
                           -\cos(\arv_{d})\cos(\cmean_{\jj})+\cos(\cmean_{d})\cos(\cmean_{\jj}))
                \notag
                \\
        &+2\sum_{d=1}^{D}\sum_{\jj=1}^{d-1}
                w_{D+d, D+\jj}(\sin(\arv_{d})\sin(\arv_{\jj})-\sin(\cmean_{d})\cos(\arv_{\jj})
                           -\sin(\arv_{d})\sin(\cmean_{\jj})+\sin(\cmean_{d})\cos(\cmean_{\jj}))
                \notag
                \\
        &+ \sum_{d=1}^{D}\sum_{\jj=1}^{D}
                w_{d,D+\jj}(\cos(\arv_{d})\sin(\arv_{\jj})-\cos(\cmean_{d})\sin(\arv_{\jj})-\cos(\arv_{d})\sin(\cmean_{\jj})+\cos(\cmean_{d})\sin(\cmean_{\jj}))
                \notag
                \\
        &+ \sum_{d=1}^{D}\sum_{\jj=1}^{D}
                w_{D+d, \jj}(\sin(\arv_{d})\cos(\arv_{\jj})-\sin(\cmean_{d})\cos(\arv_{\jj})-\sin(\arv_{d})\cos(\cmean_{\jj})+\sin(\cmean_{d})\cos(\cmean_{\jj}))
                \notag.
\end{align}

By aggregating all terms that are independent of $\arv$ and rearranging terms, Equation \eqref{eq:expanded_prod2} becomes
\begin{align}
        &\sum_{d=1}^{D}\sum_{\jj=1}^{D}
                w_{d, \jj}\cos(\arv_{d})\cos(\arv_{\jj}) + 2
                w_{d,D+\jj}\cos(\arv_{d})\sin(\arv_{\jj}) +
                w_{D+d,D+\jj}\sin(\arv_{d})\sin(\arv_{\jj}) \label{eq:aggregated}
                \\
        &-2\sum_{d=1}^{D}\sum_{\jj=1}^{D}
                w_{d,\jj}\cos(\arv_{d})\cos(\cmean_{\jj}) +
                w_{d,D+\jj}\cos(\arv_{d})\sin(\cmean_{\jj}) \notag
                \\
        &-2\sum_{d=1}^{D}\sum_{\jj=1}^{D}
                w_{D+d,D+\jj}\sin(\arv_{d})\sin(\cmean_{\jj}) +
                w_{D+d,\jj}\sin(\arv_{d})\cos(\cmean_{\jj}) \notag
\end{align}

These sums can be written in matrix notation as
\begin{align}
        \vcstd_{c}^{\top} \cos(\varv-\vcmean)
        + \vcstd_{s}^{\top} \sin(\varv-\vcmean)
        -\frac{1}{2}
        \begin{bmatrix}
          \cos(\varv) \\
          \sin(\varv)
        \end{bmatrix}^{\top}
        \begin{bmatrix}
          \matW^{cc} & \matW^{cs}\\
          (\matW^{cs})^\top & \matW^{ss}
        \end{bmatrix}
        \begin{bmatrix}
          \cos(\varv) \\
          \sin(\varv)
        \end{bmatrix}
    \label{eq:constrained_quadratic}
\end{align}
where $\cstd_{d}=\text{abs}\{z_{d}\}$ and $\cmean_{d}=\text{arg}\{z_{d}\}$ with the real and imaginary parts of $z_{d}$ such that
\begin{equation*}
  \Re\{z_{d}\} = -2\sum_{\jj=1}^{D} w_{d, \jj}\cos(\arv_{d})\cos(\cmean_{\jj}) + w_{d,D+\jj}\cos(\arv_{d})\sin(\cmean_{\jj})
\end{equation*}
and
\begin{equation*}
\Im\{z_{d}\} = -2\sum_{\jj=1}^{D} w_{D+d, D+\jj}\sin(\arv_{d})\sin(\cmean_{\jj}) + w_{D+d, \jj}\sin(\arv_{d})\cos(\cmean_{\jj}).
\end{equation*}

Therefore, Equation \eqref{eq:constrained_quadratic} imples that a multivariate Gaussian distribution under radial transformation and conditionning to the unit circle yields the log density
\begin{align}
    \log p(\varv)
        &= \const +
        \vcstd^{\top} \cos(\varv-\vcmean) \notag \\
        &-\frac{1}{2} \Big(
        \cos(\varv)^{\top}\matW^{cc}\cos(\varv) + 2
        \cos(\varv)^{\top}\matW^{cs}\sin(\varv) +
        \sin(\varv)^{\top}\matW^{ss}\sin(\varv) \Big)
        \label{eq:log_mGvM}
\end{align}
which is the log density of a multivariate Generalised von Mises distribution in overparametrised form.

To obtain the minimal number of parameters for the mGvM, Equation \eqref{eq:aggregated} can be further simplified using trigonometric identities to yield the minimal form of the mGvM distribution
\begin{align}
    \log p(\varv)
        &= \const +
    \exp
      \Big\{
          \vcstd_{1}^{\top}\cos(\varv - \vcmean_{1})
      	+ \vcstd_{2}^{\top}\cos(2 (\varv - \vcmean_{2})) \notag \\
        &
        + \frac{1}{2} \sum_{d=1}^{D}\sum_{j=1}^{D} u_{d,j} \cos(\arv_{d}-\arv_{j} - \alpha_{d,j}) + v_{d,j} \cos(\arv_{d} + \arv_{j} - \beta_{d,j})
        \Big\}.
        \label{minimal-mgvm}
\end{align}
where $\vcstd_{1}=\vcstd$ and $\vcmean_{1}=\vcmean$ given as before, while $\cstd_{2,d}=\text{abs}\{z_{d}\}$ and $\cmean_{2,d}=0.5\text{arg}\{z_{2,d}\}$ with
\begin{equation*}
  z_{d} = \frac{1}{4}(w_{d, d} - w_{D+d, D+d}) + i\frac{1}{2}(w_{d, D+d})
\end{equation*}
and the cross terms given by $u_{d,j}=\text{abs}\{z^{U}_{d,j}\}$, $\alpha_{d,j}=0.5\text{arg}\{z^{U}_{d,j}\}$, $v_{d,j}=\text{abs}\{z^{V}_{d,j}\}$, $\beta_{d,j}=0.5\text{arg}\{z^{V}_{d,j}\}$ where
\begin{align*}
  z^{U}_{d,j} &= (w_{d,j} + w_{D+d, D+j}) + i (w_{j,D+d} - w_{d, D+j})\\
  z^{V}_{d,j} &= (w_{d,j} - w_{D+d, D+j}) + i (w_{j,D+d} + w_{d, D+j}).
\end{align*}

A final point to make about the mGvM derivation is related to the distributions it generalises. \citet{gatto_generalized_2007} discussed that the GvM could be constructed by conditioning a 2D Gaussian the unit circle, but were not aware of multivariate generalisations. \citet{mardia_multivariate_2008} constructed the multivariate mvM, which we show is a submodel of the mGvM, but did not relate it to the a multivariate Gaussian nor to kernels.

\subsection{Informal argument for mGvM being the maximum entropy distribution on the hyper-torus\label{sec:maxent}}

The maximum entropy distribution $p$ subject to specified covariance and the first and second moments is the solution for the problem
\begin{align}
  \text{minimize}_{p} & \int p(\vt{x}) \log p(\vt{x}) \del \varv \\
  \text{subject to}  & \int x_{d}^{m} p(\varv) \del \arv_{d} = \alpha_{d,m}, \quad d = 1, \ldots, 2D; m = 0, \ldots, 2
\end{align}
is the multivariate Gaussian distribution. If we further add the constraints to the maximum entropy problem that the distribution must be under the unit circle, the problem becomes
\begin{align}
  \text{minimize}_{p} & \int p(\vt{x}) \log p(\vt{x}) \del \varv \\
  \text{subject to}  & \int x_{d}^{m} p(\varv) \del \arv_{d} = \alpha_{d,m}, \quad d = 1, \ldots, 2D; m = 0, \ldots, 2 \\
                     & x_{d}^{2} + x_{d+D}^{2} = 1, \quad d = 1, \ldots, D;
\end{align}
the solution of which is a multivariate Gaussian constrained to the unit hyper-torus, hence, a the mGvM distribution.

\subsection{Conditionals of the mGvM: derivation and their relationship to inference algorithms\label{sec:conditionals}}

The conditionals of the mGvM can be found by expanding the terms containing cosine of the difference and sum of two circular variables terms using sum-to-product relations. More precisely,
if we partition the indexes of mGvM distributed circular vector $\varv$ into two disjoint sets $\mathcal{A}$ and $\mathcal{B}$,
\begin{align}
    p(\varv_{\mathcal{A}}|\varv_{\mathcal{B}})
        &\propto
    \exp
      \Big\{
          \vcstd_{1,\mathcal{A}}^{\top}\cos(\varv_{\mathcal{A}} - \vcmean_{1})
          \vcstd_{1,\mathcal{B}}^{\top}\cos(\varv_{\mathcal{B}} - \vcmean_{1})
          \notag \\
          &
      	+ \vcstd_{2,\mathcal{A}}^{\top}\cos(2 (\varv_{\mathcal{A}} - \vcmean_{2})) 
      	+ \vcstd_{2,\mathcal{B}}^{\top}\cos(2 (\varv_{\mathcal{B}} - \vcmean_{2}))
        \notag \\
        &
        + \frac{1}{2} \sum_{d\in\mathcal{A}}\sum_{j\in\mathcal{A}} u_{d,j} \cos(\arv_{d}-\arv_{j} - \alpha_{d,j}) + v_{d,j} \cos(\arv_{d} + \arv_{j} - \beta_{d,j})
        \notag \\
        &
        + \frac{1}{2} \sum_{d\in\mathcal{A}}^{D}\sum_{j\in\mathcal{B}}^{D} u_{d,j} \cos(\arv_{d}- \arv_{j} - \alpha_{d,j}) + v_{d,j} \cos(\arv_{d} + \arv_{j} - \beta_{d,j})
        \notag \\
        &
        + \frac{1}{2} \sum_{d\in\mathcal{B}}^{D}\sum_{j\in\mathcal{A}}^{D} u_{d,j} \cos(\arv_{d}- \arv_{j} - \alpha_{d,j}) + v_{d,j} \cos(\arv_{d} + \arv_{j} - \beta_{d,j})
        \notag \\
        &
        + \frac{1}{2} \sum_{d\in\mathcal{B}}^{D}\sum_{j\in\mathcal{B}}^{D} u_{d,j} \cos(\arv_{d}- \arv_{j} - \alpha_{d,j}) + v_{d,j} \cos(\arv_{d} + \arv_{j} - \beta_{d,j})
      \Big\}.
\end{align}
If we consider that the variables whose indexes are in $\mathcal{B}$ are constant and note that the cross terms between variables in $\mathcal{A}$ and $\mathcal{B}$ have the functional form of $\cstd\cos(\arv_{\mathcal{A}} - \cmean)$, we can rewrite the conditional using phasor arithmetic as
\begin{align}
    p(\varv_{\mathcal{A}}|\varv_{\mathcal{B}})
        &\propto
    \exp
      \Big\{
          \tilde{\vcstd}_{1}^{\top}\cos(\varv_{\mathcal{A}} - \tilde{\vcmean_{1}})
      	+ \vcstd_{2,\mathcal{A}}^{\top}\cos(2 (\varv_{\mathcal{A}} - \vcmean_{2}))
        \notag \\
        &
        + \frac{1}{2} \sum_{d\in\mathcal{A}}\sum_{j\in\mathcal{A}} u_{d,j} \cos(\arv_{d}-\arv_{j} - \alpha_{d,j}) + v_{d,j} \cos(\arv_{d} + \arv_{j} - \beta_{d,j})
\end{align}
where $\tilde{\cstd}_{1,d}=\text{abs}(z_d)$, $\tilde{\vcmean}_{1,d}=\text{arg}(z_d)$ and
\begin{align*}
  \Re(z_d) &= \cstd_{1,d}\cos(\cmean_{1,d}) + \sum_{j=1} u_{d,j}\cos(\arv_{j} - \alpha_{d,j}) - v_{d,j}\cos(\arv_{j} - \beta_{d,j})\\
  \Im(z_d) &= \cstd_{1,d}\sin(\cmean_{1,d}) + \sum_{j=1} u_{d,j}\sin(\arv_{j} - \alpha_{d,j}) - v_{d,j}\sin(\arv_{j} - \beta_{d,j})
\end{align*}
with $\Re(z)$ denoting the real part of $z$ and $\Im(z)$ denoting the imaginary part of $z$.

In the particular case of the unidimensional conditional, the covariance term
\begin{equation*}
  + \frac{1}{2} \sum_{d\in\mathcal{A}}\sum_{j\in\mathcal{A}} u_{d,j} \cos(\arv_{d}-\arv_{j} - \alpha_{d,j}) + v_{d,j} \cos(\arv_{d} + \arv_{j} - \beta_{d,j})
\end{equation*}
will vanish as the diagonals of the parameter matrices $\mat{U}$ and $\mat{V}$ are zero.

\subsubsection{Unidimensional conditionals and Gibbs sampling}

When the set $\mathcal{A}$ contains a single index, the expressions in the previous section define how to obtain all unidimensional conditionals of the mGvM. These one-dimensional conditionals are the same mentioned in the main paper for the Gibbs sampling, for which the parametric dependencies can be explicitly written as
\begin{align}
  \tilde{\cstd}_{1,d} = \text{abs}(z_d) & \tilde{\cmean}_{1,d} = \text{arg}(z_d)
\end{align}
where $z_{d}=$ is given by
\begin{align*}
  \Re(z_d) &= \cstd_{1,d}\cos(\cmean_{1,d}) + \sum_{j=1} u_{d,j}\cos(\arv_{j} - \alpha_{d,j}) - v_{d,j}\cos(\arv_{j} - \beta_{d,j})\\
  \Im(z_d) &= \cstd_{1,d}\sin(\cmean_{1,d}) + \sum_{j=1} u_{d,j}\sin(\arv_{j} - \alpha_{d,j}) - v_{d,j}\sin(\arv_{j} - \beta_{d,j})
\end{align*}
with $\Re(z)$ denoting the real part of $z$ and $\Im(z)$ denoting the imaginary part of $z$.

\subsubsection{Unidimensional conditionals and mean field variational approximation}

To find the approximation $q(\arv|\rho)$ for a true posterior $p(\arv|\psi, \theta)$ that minimisesthe Kullback-Leiber divergence from $q$ to $p$ under the variational free energy framework~\cite{jordan1999}, can equivalently maximise the variational free energy $\free(\q,\params_{\p})$ by noting

\begin{align*}
\text{KL}(q(\varv)||p(\varv|\vt{\psi}))
    &= \int q(\varv|\rho)\log \frac{\q(\varv|\rho)}{p(\varv|\vt{\psi},\theta)}\del \varv \\
    &=-\int q(\varv|\rho)\log p(\arv|\vt{\psi},\theta)\del \arv +\int q(\varv|\rho)\log q(\varv|\rho) \del \varv\\
    &=-\langle\log \p(\varv,\vt{\psi}|\theta)\rangle_{\q(\varv|\rho)} +\log\p(\vt{\psi}|\theta) - \mathcal{H}(q) \\
    &= \log \p(\vt{\psi}|\theta) - \free(q, \theta, \rho).
\end{align*}
By assuming a fully factored form for the distribution $q$, i.e., $q(\varv)=\prod_{d=1}^{D} q_{d}(\arv_{d})$, we can use calculus of variations to obtain analytically the functional form of the distributions $q_d$, that is
\begin{equation}
    \frac{\delta}{\delta \q} \free(\q,\params_{\p}) - \lambda \left(\int \q(\varv)\del \varv - 1 \right) = 0
\end{equation}
which implies
\begin{align}
    \frac{\delta}{\delta \q_{\ell}}
    \left[
        \expect{\log \p(\arv,\psi)}_{\prod_{\dd}^{\DD}\q_{\dd}(\arv_{\dd})}
        -\int \prod_{\dd=1}^{\DD} \q_{\dd}(\arv_{\dd})
              \left(\sum_{\dd=1}^{\DD}\log \q_{\dd}(\arv_{\dd})\right)
         \del \varv
        - \lambda \sum_{\dd=1}^{\DD}
                \int \q_{\dd}(\arv_{\dd}) d\arv_{\dd}
    \right] = 0
\end{align}
and leads to the factors approximation
\begin{equation}
    \q_{d}(\arv_{d}) = \frac{1}{\exp(\lambda+1)}\exp
        \left\{
            \expect{\log \p(\arv,\psi)}_{\prod_{\neq d}^{D}\q_{\neq d}(\arv_{\neq d})}
        \right\}
\label{eq:meanFieldEquation}
\end{equation}
resulting in the set of distributions known as mean field approximation. This equation when applied to the mGvM yields GvM distributions 
\begin{align}
  q(\arv_{d}|\varv_{\neq d}) = \gvm(\arv_{d};
  \bar{\cstd}_{1,d}(\vcstd,\vcmean, \langle e^{i\varv_{\neq d}}\rangle_{q_{\neq d}}, \langle e^{i \varv_{\neq d}} \rangle_{q_{\neq d}}),\cstd_{2,d},
  \bar{\cmean}_{1,d}(\vcstd,\vcmean, \langle e^{i\varv_{\neq d}}\rangle_{q_{\neq d}}, \langle e^{i \varv_{\neq d}} \rangle_{q_{\neq d}}), \cmean_{2,d})
\end{align}
where $\langle e^{i n \varv}\rangle= \langle \cos(n \varv)\rangle + i\langle \sin(n \varv)\rangle $, $\bar{\cstd}_{1,d} = \text{abs}(z_d)$ and $\bar{\cmean}_{1,d} = \text{arg}(z_d)$ with $z_{d}=$ given in overparametrised form by
\begin{align*}
  \Re(z_d) &= \cstd_{d}\cos(\cmean_{1,d}) -\frac{1}{2} \left\langle
        \begin{bmatrix}
          \cos(\varv_{\neq d}) \\
          \sin(\varv_{\neq d})
        \end{bmatrix}^{\top}
        \begin{bmatrix}
          \matW^{cc}_{\neq d, d} & \matW^{cs}_{\neq d, d}\\
          (\matW^{cs})_{\neq d, d}^\top & \matW^{ss}_{\neq d, d}
        \end{bmatrix}
        \begin{bmatrix}
          \cos(\varv_d) \\
          \sin(\varv_d)
        \end{bmatrix}
        \right\rangle_{q_{\neq \varv_d}}
        \\
  \Im(z_d) &= \cstd_{d}\sin(\cmean_{1,d}) -\frac{1}{2} \left\langle
        \begin{bmatrix}
          \cos(\varv_{\neq d}) \\
          \sin(\varv_{\neq d})
        \end{bmatrix}^{\top}
        \begin{bmatrix}
          \matW^{cc}_{\neq d, d} & \matW^{cs}_{\neq d, d}\\
          (\matW^{cs})_{\neq d, d}^\top & \matW^{ss}_{\neq d, d}
        \end{bmatrix}
        \begin{bmatrix}
          \cos(\varv_d) \\
          \sin(\varv_d)
        \end{bmatrix}
        \right\rangle_{q_{\neq \varv_d}}
\end{align*}
with $\Re(z)$ denoting the real part of $z$ and $\Im(z)$ denoting the imaginary part of $z$.

\subsection{Higher order mGvM\label{sec:higher}}

As with the higher order GvM, the mGvM can also be expanded to include $T$ cosine harmonics if the Gaussian genesis is cast aside. In this case, a mGvM of order $T$ can be defined as
\begin{align}
	\mgvm_{T}(\varv; \vcmean_{1:T}, \vcstd_{1:T}, \mat{U}, \mat{V}, \mat{\alpha}, \mat{\beta})
    &\propto 
    \exp
      \Big\{
          \sum_{t=1}^{T} \vcstd_{t}^{\top}\cos(t (\varv - \vcmean_{t})) \notag \\
        &
        + \frac{1}{2} \sum_{i=1}^{D}\sum_{j=1}^{D} u_{i,j} \cos(\arv_{i}-\arv_{j} - \alpha_{i,j}) + v_{i,j} \cos(\arv_{i}+\arv_{j} - \beta{i,j})
      \Big\}
      \label{eq:mgvm_2}
\end{align}
which is a distribution whose conditionals allow up to $T$ modes, but bears the same correlation correlation structure of the `standard' order 2 mGvM.$\mathcal{A}$ is a single index, the 

\subsection{Assumptions over the precision matrix of the Gaussian that leads to a mvM}

In this section the mvM is derived by conditioning a 4-dimensional multivariate Gaussian to highlight the assumptions made regarding the precision matrix of the multivariate Gaussian. We take the 4D Gaussian to be zero mean without loss of generality and, after applying the polar variable transformation and constraining the radial components to unity we obtain the distribution
\begin{align}
  p(\phi_1, \phi_2)
  &\propto \exp
     \left\{
       -\frac{1}{2}
       \begin{bmatrix}
         \cos(\phi_1)\\
         \cos(\phi_2)\\
         \sin(\phi_1)\\
         \sin(\phi_2)
       \end{bmatrix}^{\top}
       \begin{bmatrix}
         a_{1,1} & a_{1,2} & a_{1,3} & a_{1,4} \\
         a_{1,2} & a_{2,2} & a_{2,3} & a_{2,4} \\
         a_{1,3} & a_{2,3} & a_{3,3} & a_{3,4} \\
         a_{1,4} & a_{2,4} & a_{3,4} & a_{4,4} \\         
       \end{bmatrix}
       \begin{bmatrix}
         \cos(\phi_1)\\
         \cos(\phi_2)\\
         \sin(\phi_1)\\
         \sin(\phi_2)
       \end{bmatrix}
     \right\}
\end{align}
which can be expanded into
\begin{align}
  p(\phi_1, \phi_2)
  &\propto \exp
     \Big\{
        - \frac{1}{2}
        (
        a_{1,1} \cos(\phi_1)^2 + a_{2,2} \cos(\phi_2)^2 + a_{3,3} \sin(\phi_1)^2 + a_{4,4} \sin(\phi_2)^2 + \notag \\
        & 2 a_{1,2} \cos(\phi_1) \cos(\phi_2) + 2 a_{1,3} \cos(\phi_1) \sin(\phi_1) + 2 a_{1,4} \cos(\phi_1) \sin(\phi_2) + \notag \\
        & 2 a_{2,3} \cos(\phi_2) \sin(\phi_1) + 2 a_{2,4} \cos(\phi_2) \sin(\phi_2) + 2 a_{3,4} \sin(\phi_1) \sin(\phi_2)
        )
     \Big\}.
     \notag
\end{align}
Using the fundamental identity and double angle formulas, we can rewrite the last Equation as
\begin{align}
  p(\phi_1, \phi_2)
  &\propto \exp
     \Big\{
        - \frac{1}{2}
        (
        a_{1,1} \cos(\phi_1)^2 + a_{2,2} \cos(\phi_2)^2 + a_{3,3} (1 - \cos(\phi_1)^2) + a_{4,4} (1 - \cos(\phi_2)^2) + \notag \\
        & 2 a_{1,2} \cos(\phi_1) \cos(\phi_2) + a_{1,3} \sin(2 \phi_1) + 
          2 a_{1,4} \cos(\phi_1) \sin(\phi_2) + 2 a_{2,3} \cos(\phi_2) \sin(\phi_1) + \notag \\
        & a_{2,4} \sin(2 \phi_2) + 2 a_{3,4} \sin(\phi_1) \sin(\phi_2)) 
        )
     \Big\} \notag
\end{align}

Further simplifications arise from
\begin{align}
  p(\phi_1, \phi_2)
  &\propto \exp
     \Big\{
        - \frac{1}{2}
        (
        (a_{1,1} - a_{3,3}) \cos(2 \phi_1 - 2 \nu_1) + (a_{2,2} - a_{4,4}) \cos(2 \phi_2 - 2 \nu_2) + \notag \\
        & 2 a_{1,2} \cos(\phi_1) \cos(\phi_2) +  2 a_{1,4} \cos(\phi_1) \sin(\phi_2) +
          2 a_{2,3} \cos(\phi_2) \sin(\phi_1) +  \notag \\
        & 2 a_{3,4} \sin(\phi_1) \sin(\phi_2) + a_{1,3} \sin(2 \phi_1) + a_{2,4} \sin(2 \phi_2) )
     \Big\}. \notag
\end{align}
The product of sine and cosines can be also translated using product-to-sum formulas
\begin{align}
  p(\phi_1, \phi_2)
  &\propto \exp
     \Big\{
        - \frac{1}{2}
        (
        (a_{1,1} - a_{3,3}) \cos(2 \phi_1 - 2 \nu_1) + (a_{2,2} - a_{4,4}) \cos(2 \phi_2 - 2 \nu_2) + \notag \\
        & a_{1,2} \cos(\phi_1 - \phi_2) + a_{1,2} \cos(\phi_1 + \phi_2) + 
          a_{1,4} \sin(\phi_1 + \phi_2) - a_{1,4} \sin(\phi_1 - \phi_2) + \notag \\
        & a_{2,3} \sin(\phi_1 + \phi_2) + a_{2,3} \sin(\phi_1 - \phi_2) + 
          a_{3,4} \cos(\phi_1 - \phi_2) - a_{3,4} \cos(\phi_1 + \phi_2) + \notag \\
        & a_{1,3} \sin(2 \phi_1) + a_{2,4} \sin(2 \phi_2))
     \Big\} \notag
\end{align}
grouping similar terms
\begin{align}
  p(\phi_1, \phi_2)
  &\propto \exp
     \Big\{
        - \frac{1}{2}
        (
        (a_{1,1} - a_{3,3}) \cos(2 \phi_1 - 2 \nu_1) + (a_{2,2} - a_{4,4}) \cos(2 \phi_2 - 2 \nu_2) + \notag \\
        & a_{1,3} \sin(2 \phi_1) + a_{2,4} \sin(2 \phi_2) + \notag \\
        & (a_{1,2} + a_{3,4}) \cos(\phi_1 - \phi_2) + (a_{1,2} - a_{3,4}) \cos(\phi_1 + \phi_2) + \notag \\
        & (a_{2,3} - a_{1,4}) \sin(\phi_1 - \phi_2) + (a_{1,4} + a_{2,3}) \sin(\phi_1 + \phi_2))
     \Big\} \notag
\end{align}
Therefore, we can conclude that since the mvM has only the term $\sin(\phi_1 - \phi_2)$ from the equation above that $a_{1,1}=a_{3,3}$, $a_{2,2}=a_{4,4}$, $a_{1,3}=a_{2,4}=a_{1,2}=a_{3,4}=0$ and $a_{1,4}=-a_{3,2}$. This leads to the precision matrix having the sparsity pattern
\begin{align}
  \begin{bmatrix}
    a_{1,1} & 0       & 0       & a_{1,4} \\
    0 &  a_{2,2} & -a_{1,4} & 0      \\
    0 & -a_{1,4} &  a_{1,1} & 0      \\
    a_{1,4} & 0       & 0       & a_{2,2} \\
  \end{bmatrix}.
  \notag
\end{align}

\subsection{Dimensionality reduction with the mGvM and Probabilistic Principal Component Analysis\label{sec:CPCA}}
In this section we discuss in greater detail dimensionality reduction with the Multivariate Generalised von Mises and its relationship to Probabilistic Principal Component Analysis (PPCA) from \cite{tipping_probabilistic_1999}.

The PCA model is defined as
\begin{equation}
    \begin{aligned}
    p(\vxx) &= \gaussian(\vxx; 0, \id)\\
    p(\vyy|\vxx) &= \gaussian(\vyy; \matW\vxx, \stdev^{2} \id) 
    \end{aligned}
\end{equation}
where $\matW$ is a matrix that encodes the linear mapping between hidden components
$\vxx \in \reals^{D}$ and data $\vyy \in \reals^{\MM}$, with $\MM > D$.

If we impose that each of the latent components $x_{d}$ is sinusoidal and may be parametrised by a hidden angle $\arv_{d}$ plus a phase shift $\varphi_{d}$, we obtain the model
\begin{equation}
    \begin{aligned}
        p(\arv_{d}) &= \gvm(\arv; \cstd_{1, d}, \cstd_{2, d},
                                     \cmean_{1, d}, \cmean_{2, d})\\
        p(x_{d}|\arv_{d}) &= \delta(x_{d} - \sin(\arv_{d} + \varphi_{d}))\\
        p(\vyy|\vxx) &= \gaussian(\vyy; \matW\vxx, \stdev^{2} \id) 
    \end{aligned}
\end{equation}

To obtain the relation directly between the data and the hidden angle, we integrate out the latent components $\vxx$
\begin{equation}
        p(\vyy|\varv) = \int \delta(\vxx - \sin(\varv + \vec{\varphi}))
            \gaussian(\vyy; \matW\vxx, \stdev^{2} \id) d\vxx
\end{equation}
which results in the model used in the mGvM dimensionality reduction application.

Alternatively, it is also possible to show the limiting behaviour of the model arising fromthe mGvM dimensionality reduction application becomes the PCA model, for mean angles $\vec{\cmean} \rightarrow 0$ and high concentration parameters. In this regime, the small angle approximation
\begin{equation}
        \sin\arv \approx \arv, \quad \cos\arv \approx 0
\end{equation}
is valid and leads to the Generalised Von Mises priors simplification to
\begin{equation}
    \begin{aligned}
        p(\arv) &\propto \exp\left\{
                                        \cstd_{1}\cos(\arv - \cmean_{1}) 
                                        + \cstd_{2}\cos(2(\arv - \cmean_{2}))
                                    \right\}\\
                    &\propto \exp\left\{
                                        -\cstd_{1}\cos(\cmean_{2})\arv^{2}
                                        + (\cstd_{1}\sin(\cmean_{1})
                                           + 2 \cstd_{2}\sin(2\cmean_{2}))\arv
                                    \right\}\\
                    &\propto \exp\left\{
                                    -\cstd_{1}\cos(\cmean_{2})
                                    \left[
                                        \arv
                                        - \frac{\cstd_{1}\sin(\cmean_{1})
                                                + 2 \cstd_{2}\sin(2\cmean_{2})}{2\cstd_{1}\cos(\cmean_{2})}
                                        \right]^{2}
                                    \right\}
    \end{aligned}  
\end{equation}
which is proportional to a Gaussian distribution and shows that under the small angle regime, the coefficient matrix $\matA$ is a good approximation for $\matW$ and the model collapses to PCA.

\begin{figure}[tbp]
\centering
\includegraphics[width=0.75\textwidth]{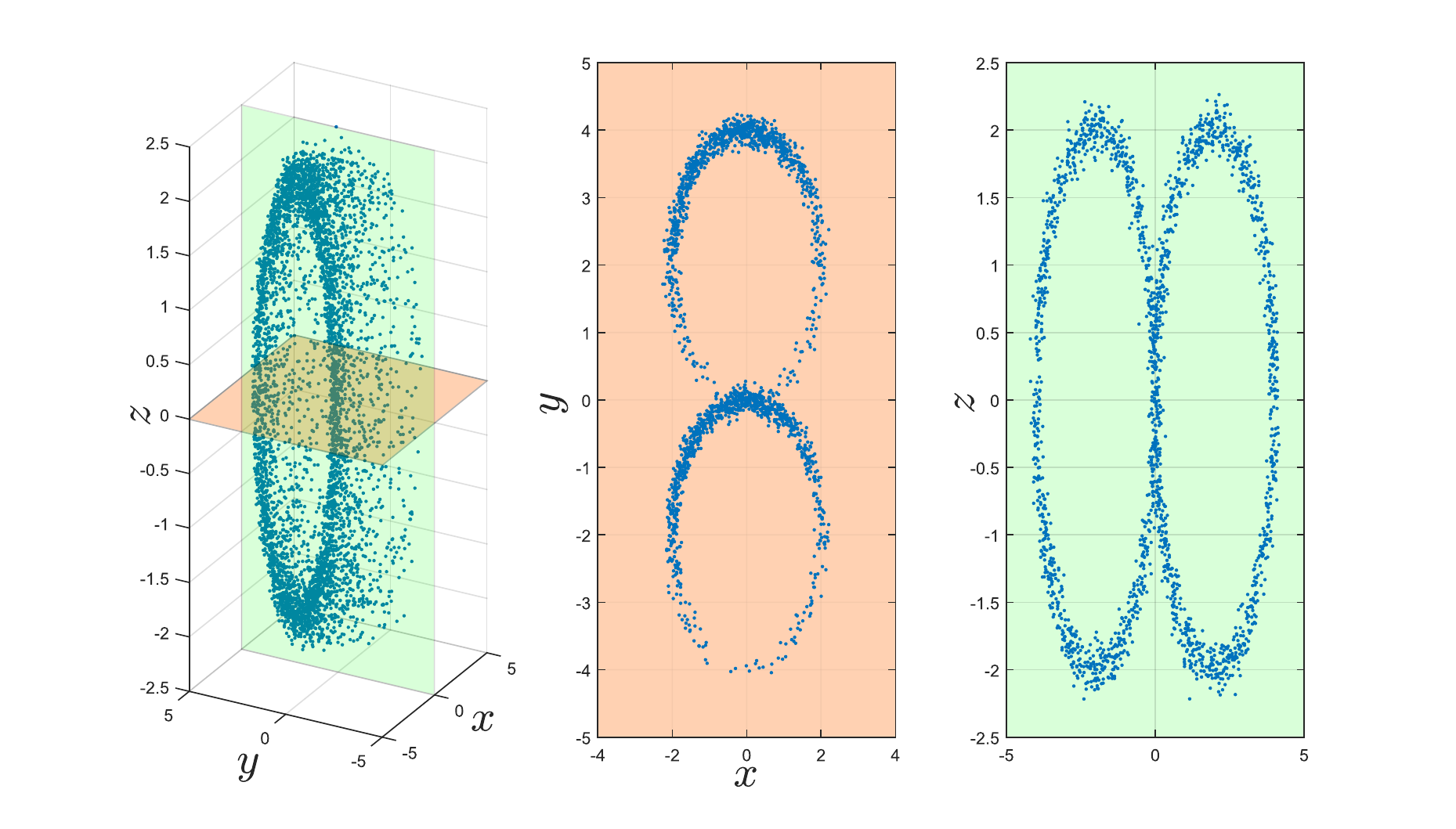}
\caption{Plots of the model $x = 2 \cos{\arv_{1}} + \error$, $y = 2 \sin{\arv_{1}} + 2 \cos{\arv_{2}} + \error$, $z = 2 \sin{\arv_{2}} + \error$ where $\arv_{1} \sim \vm(50, \pi/2)$ is a peaked von Mises distribution, $\arv_{2} \sim \vm(0.1, 0)$ is an almost-uniform von Mises distribution and the noise is $\error \sim \gaussian(0, 0.01)$ to exemplify a 3-dimensional Cartesian data set as a function of a 2-dimensional 
angular space: plot of samples from the model (left), samples on the $z = 0$ plane, which is equivalent to fixing $\arv_{2}=\pm \pi$ (middle), samples on the $x = 0$ plane, which is equivalent to fixing $\arv_{1}=\pm \pi/2$ (right).}
\label{fig:doughnut}
\end{figure}

Another connection between the dimensionality reduction with the mGvM and PCA may be established geometrically. While PCA describes the data in terms of hidden hyperplanes, the lower dimensional description of the data with mGvM occurs in terms of hidden tori, as illustrated in \cref{fig:doughnut}.
The effect of priors in this systems is also highlighted by \cref{fig:doughnut}. The mean angle and concentration of each prior impacts the distribution of mass along the direction of the angular component on the hyper-torus. High concentration values on the prior leads to dense regions around the mean angle, as presented in the middle graph of \cref{fig:doughnut} while low concentration leads to uniform mass distribution, shown in the right graph of \cref{fig:doughnut}.

An analogy often used to describe this shape of the data in the PCA's hidden space is a ``fuzzy pancake'', as the Gaussian noise induces the shape irregularity (``fuzzyness''), of the hidden plane (``pancake''). Likewise, for dimensionality reduction with the mGvM the corresponding analogy would be a ``fuzzy doughnut'', as the Gaussian noise also incur in irregularities over the surface of a ``doughnut'', which bears similar shape to a torus.

\subsection{Supporting graphs and analysis for the experiments}

\subsubsection{Regression experiments}

\begin{figure*}[tbp]
  \begin{tabular}{c@{\hspace{8mm}}c@{\hspace{8mm}}c}
    \raisebox{-0.5\height}{\includegraphics[width=0.35\textwidth]{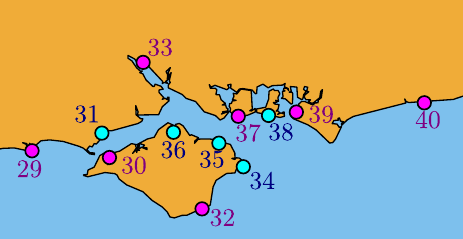}} &
    \raisebox{-0.5\height}{\includegraphics[width=0.25\textwidth]{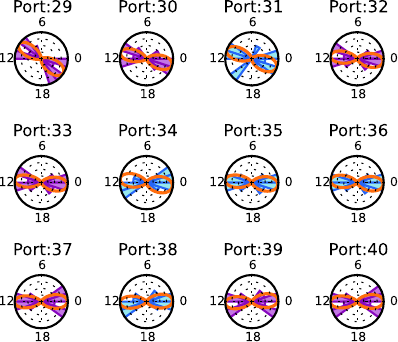}} &
    \raisebox{-0.5\height}{\includegraphics[width=0.25\textwidth]{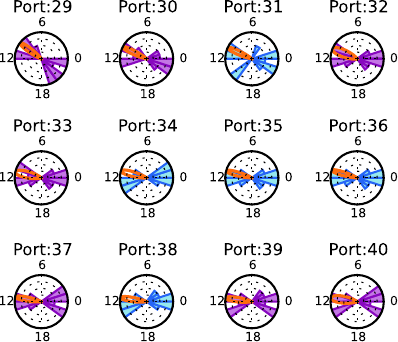}} \\[16mm]
    \small{Port locations} & \small{mGvM} & \small{2D-GP}
  \end{tabular}
    \caption{Tide time predictions on the UK coast: port location for a subset of the dataset (left), mGvM fit (left) and 2D-GP (right). The ports whose data was supplied for training are displayed in magenta (darker) rose diagrams whereas the ports held out for prediction are displayed in cyan (lighter). The regression model predictions are given as orange lines. Best visualised in colour.}
    \label{fig:tides}
\end{figure*}

The plots in \cref{fig:1d_regression,fig:tides} help us understand the some reasons why the mGvM provides better regression performance than the other models considered. The mGvM is able to accurately infer where the underlying function wraps, and provides a reasonable estimate for both the expected value of the underlying function and variance on the unit circle. The 1D-GP cannot account for the angular equivalences, therefore, it has assign this phenomenon to noise resulting in flat predictions as shown in \cref{fig:1d_regression}. While the 2D-GP is able to cope with wrapping, it learns a different lengthscale parameter. Furthermore, the 2D-GP cannot learn bimodal errors which can be accounted for by the mGvM as shown in \cref{fig:tides}.

\begin{figure*}[tbp]
  \centerline{
    \includegraphics[width=0.33\textwidth]{uni_model0_mgvm.pdf}
    \includegraphics[width=0.33\textwidth]{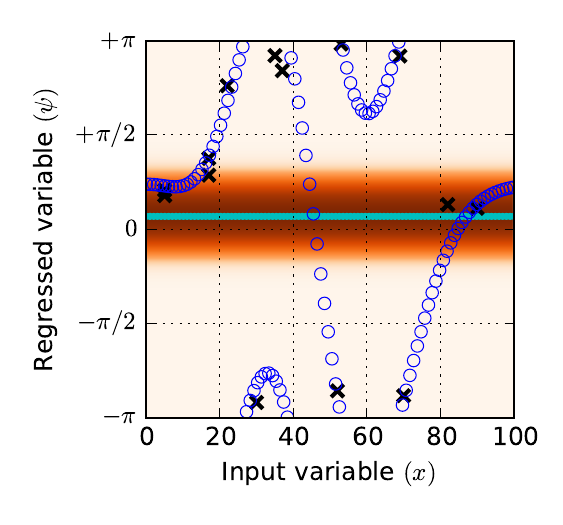}
    \includegraphics[width=0.33\textwidth]{uni_sincos_gp.pdf}
  }
  \caption{Regression on a one-dimensional a synthetic data set using the mGvM (left), 1D GP (center) and 2D GP (right): data points are represented as balck crosses, the true function with circles and model predictions in solid dots. Best visualised in colour.}
  \label{fig:1d_regression}
\end{figure*}

\subsubsection{Dimensionality reduction experiments}

\begin{figure}[tbp]
\vskip 0.2in
\begin{center}
\centerline{\includegraphics[width=0.9\columnwidth]{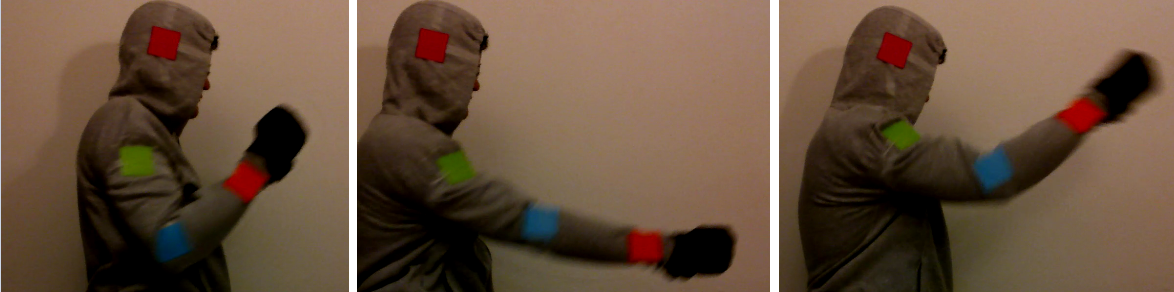}}
\caption{Capturing 2D motion: the datasets was generated by recording the motion of a subject with markers on its body then using a colour threshold algorithm and taking the location of the centre of mass of the filtered region.}
\label{fig:mocap_xp}
\end{center}
\vskip -0.2in
\end{figure}

In this section, we provide additional experiments and noise values for the average signal-to-noise ratio for motion capture and the simulation of motion capture of a robot arm. In the motion capture data sets, we applied a colour filter to the resulting images to isolate each marker and then the marker position was found by calculating the centre of mass of each marker as shown in Figure \ref{fig:mocap_xp}.

\begin{figure*}[tbp]
\begin{center}
\centerline{
  \includegraphics[width=0.33\textwidth]{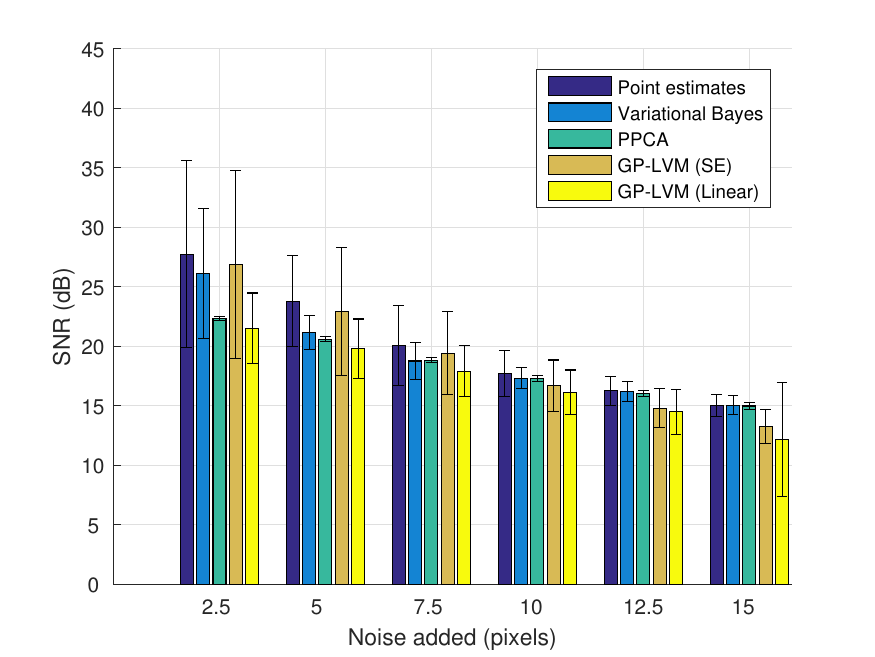}
  \includegraphics[width=0.33\textwidth]{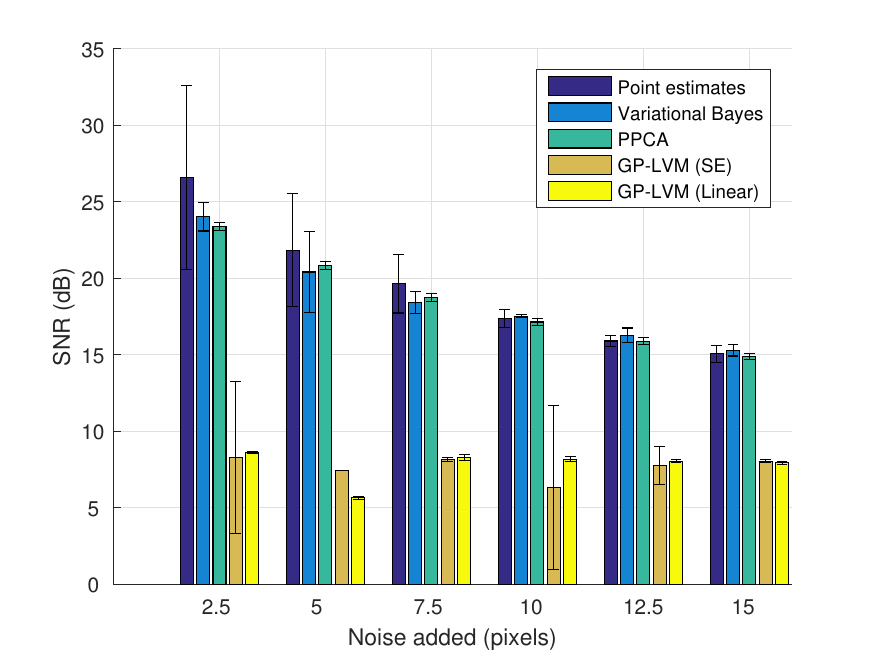}
  \includegraphics[width=0.33\textwidth]{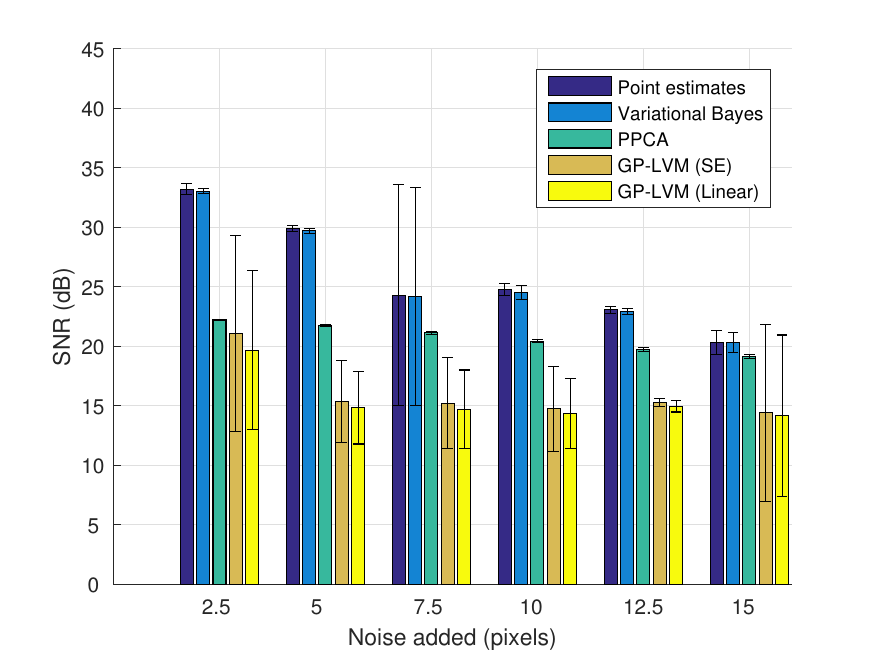}
}
\caption{Signal-to-noise ratio with 3 standard deviations for the latent variable modelling datasets: filmed subject running(left), fishing (middle) and synthetic dataset (right).}
\label{fig:mocap_results}
\end{center}
\end{figure*}

Additional experiments are given in \cref{fig:mocap_results}. The conclusions and discussion of these experimental results mirror the discussions presented in the main paper.


\end{document}